\documentclass[11pt]{article}

\usepackage{epsfig,amsmath,latexsym,amssymb}
\usepackage{graphicx}
\usepackage{lscape}
\usepackage{picture, eso-pic, tikz} 
\usepackage{dsfont}
\usepackage{listings}
\usepackage{changes}
\usepackage{hyperref}

\def\R{{\mathbb R}}  
\def\N{{\mathbb N}}  
\def\p{{\mathbb P}}  
\def\E{{\mathbb E}}  %

\newcommand{\Remm}[1]{}
\newtheorem{theo}{Theorem}[section]

\newtheorem{proposition}[theo]{Proposition}

\newtheorem{definition}[theo]{Definition}
\newtheorem{model ass}[theo]{Model Assumptions}

\newtheorem{example}[theo]{Example}

\newtheorem{remark}[theo]{Remark}

\makeatletter
\def\thm@space@setup{%
  \thm@preskip=8pt plus 2pt minus 4pt
  \thm@postskip=\thm@preskip
}
\makeatother

\def\EndExample{\hfill {\scriptsize $\blacksquare$}}
\numberwithin{equation}{section}

\definecolor{MyGray}{rgb}{0.92,0.92,0.92}


\newcommand{\bl}[1]{\textcolor{blue}{{#1}}}

\definecolor{British racing}{rgb}{0.0, 0.5, 0.0}
\def\bx{\boldsymbol{x}}
\def\be{\boldsymbol{e}}

\def\bX{\boldsymbol{X}}
\def\b0{\boldsymbol{0}}

\def\b0{\boldsymbol{0}}

\def\bphi{\boldsymbol{\phi}}

\title{Tree-like Pairwise Interaction Networks}
\author{Ronald Richman\footnote{InsureAI and University of the Witwatersrand, ronaldrichman@gmail.com}
\and Salvatore Scognamiglio\footnote{Department of Management and Quantitative Sciences, University of Naples ``Parthenope",\newline salvatore.scognamiglio@uniparthenope.it}
\and Mario V.~W\"uthrich\footnote{Department of Mathematics, ETH Zurich,
mario.wuethrich@math.ethz.ch}}

\date{\today}
\begin{document}

\maketitle

\begin{abstract}
Modeling feature interactions in tabular data remains a key challenge in predictive modeling, for example, as used for insurance pricing. This paper proposes the Tree-like Pairwise Interaction Network (PIN), a novel neural network architecture that explicitly captures pairwise feature interactions through a shared feed-forward neural network architecture that mimics the structure of decision trees. PIN enables intrinsic interpretability by design, allowing for direct inspection of interaction effects. Moreover, it allows for efficient
SHapley's Additive exPlanation (SHAP) computations because it only involves pairwise interactions. We highlight connections between PIN and established models such as GA²Ms, gradient boosting machines, and graph neural networks. Empirical results on the popular French motor insurance dataset show that PIN outperforms both traditional and modern neural networks benchmarks in predictive accuracy, while also providing insight into how features interact with each another and how they contribute to the predictions.

\bigskip

\end{abstract}

\section{Introduction}
Understanding and modeling the interactions among input features represents a core challenge in predictive modeling on tabular data. These interactions refer to how multiple variables jointly influence the target outcome in ways that go beyond their individual, isolated effects. In many real-world cases, the relationship between predictors and the response variable is not simply additive and may involve complex interdependencies. This is particularly relevant in insurance pricing, where factors such as driver age, location, and driving behavior can interact in non-obvious ways to affect risk assessment and premium calculation. Overlooking or misspecifying such interactions can lead to suboptimal models, resulting in price distortions and potentially biased interpretations.

Traditionally, the insurance industry has relied on Generalized Linear Models (GLMs) due to their interpretability, strong statistical foundation and also for computational reasons; we refer to Nelder--Wedderburn \cite{NelderWedderburn1972} and McCullagh--Nelder \cite{MN}. In the GLM framework, the expected value of the response variable is modeled as a linear combination of predictors, linked to the response through a specified link function. While GLMs allow for the estimation of individual covariate effects, they offer limited capability to capture interactions among variables. In fact, interactions can be integrated, but each one needs to be engineered manually which clearly limits their scope and the model complexity and accuracy.

Generalized Additive Models (GAMs) were introduced in Hastie--Tibshirani \cite{HT1} as an extension of GLMs relaxing the assumption of linearity. They allow each predictor to have a nonlinear effect on the response variable. This is accomplished by replacing the original covariates with smooth functions (typically splines) applied individually to each input variable. GAMs have often been used in the context of actuarial science as a foundation for developing more sophisticated methods to solve actuarial problems; notable examples include
Chang et al.~\cite{ChangGaoShi2024} and Maillart--Robert \cite{maillart2024gam}.
Building upon GAMs, Generalized Additive Models with Pairwise Interactions (GA²Ms) extend the framework by incorporating a selected set of pairwise feature interactions, see Lou et al.\ \cite{Lou}. These additional components enable the model to capture the joint effects of feature pairs on the response variable. Similarly to interactions in GLMs, GA²Ms typically require manual identification and specification of interaction terms, which demands domain expertise and time consuming feature engineering. This reliance on expert judgement can introduce subjectivity and biases, and it may make the model sensitive to the modeler’s choices, potentially affecting both predictive performance and generalization capability.

Deep neural networks (DNNs) have attracted a growing attention in actuarial science as powerful tools for predictive modeling; for an extended discussion see Richman \cite{Richman2021a, Richman2021b}. These models can be viewed as an extension of GLMs that allow for the modeling of nonlinear relationships; see Chapter 5 of W\"uthrich et al.\ \cite{AITools}. In particular, based on a successful training algorithm, DNNs perform automated feature engineering. Through the mechanism of representation learning, DNNs transform original input features into new, higher-level representations via multiple layers of nonlinear functions, which are optimized to predict the response variable. This hierarchical structure allows the model to automatically construct features that capture higher-order interactions among the original covariates.
Despite their strong predictive performance, DNNs present significant challenges, particularly in identifying, quantifying and interpreting feature interactions. The complexity of the learned representations makes it difficult to understand how input variables jointly contribute to the model's predictions. For these reasons, DNNs are sometimes called black box models.

Modern machine learning literature grapples with the challenge of effectively modeling and interpreting feature interactions within neural networks. Song et al.\ \cite{song2019autoint} address this issue by proposing a specialized DNN architecture designed to automatically learn feature interactions in presence of tabular data. The model employs a multi-head self-attention mechanism, inspired by the Transformer architecture of Vaswani et al.\ \cite{Vaswani}, to explicitly capture interactions between features, thereby eliminating the need for manual feature engineering. Tsang et al. \ \cite{tsang2018nid} introduce a method for detecting statistical interactions in trained DNNs by analyzing the  values of the learned weights. Li et al.\ \cite{li2019fignn} proposes modeling categorical feature interactions by representing covariates using a graph, where each node corresponds to a distinct feature field. Interactions among features are effectively captured through the propagation mechanisms of a Graph Neural Network (GNN). Enouen--Liu \cite{EnouenLiu2022} develop a DNNs-based approach that extends classical GAMs by introducing an additional component to capture higher-order feature interactions. Havrylenko--Heger \cite{Havrylenko} revisit the CANN proposal of W\"uthrich--Merz \cite{CANN} to identify the most significant interacting pairs missing in a GLM.


The contribution of this paper is to introduce the Tree-like Pairwise Interaction Network (PIN), a novel neural network architecture designed to model and identify pairwise feature interactions in tabular data. The PIN architecture embeds each input feature into a learned latent space and then explicitly models all pairwise interactions through a shared feed-forward neural network. Sharing this feed-forward neural network implies an efficient way of dealing with model parameters, and to still allow for variability between the different interactions, each interaction makes use of a dedicated set of learnable parameters -- playing a similar role to the Classification (CLS) token used in Devlin et al.~\cite{Devlin} -- which is aimed at modeling the dependence structure between each pair of features. The output of each interaction is passed through a centered hard sigmoid activation function that mimics the discrete partitioning behavior of decision trees, yet in a continuous and differentiable form. This results in tree-like structures (using binary splits), and this model allows for direct inspection of interaction terms allowing  for model explainability. 
The proposed PIN also exhibits connections with several well-established machine learning models, including gradient boosting, GA²M, and graph neural networks. These relationships are explored in detail throughout the manuscript.
We assess the performance of the Tree-like PIN on the widely studied French motor third-party liability claims frequency dataset. Our experiments demonstrate that the PIN architecture delivers strong predictive accuracy in terms of Poisson losses, outperforming not only classical benchmarks but also more recent neural network architectures, such as the Credibility Transformer introduced by Richman et al.~\cite{RSW}. Moreover,  unlike most other network architectures, the PIN  architecture allows for an efficient computation of SHapley's Additive exPlanations (SHAP). The crucial point is that the PIN architecture only involves pairwise interactions, and it has been shown in Lundberg \cite{Lundberg2018} and Mayer--W\"uthrich \cite[Proposition 3.1]{MayerW} that this allows for a very efficient computation of the Shapley values using paired permutation SHAP sampling.

\medskip

Concluding this introduction: Our novel architecture of the tree-like PIN provides superior performance on tabular data over currently used network architectures, and at the same time it provides explainability. We believe that this makes the tree-like PIN an interesting proposal for actuarial pricing.

\medskip

The rest of this paper is organized as follows. In Section \ref{sec:PIN}, we introduce the PIN architecture, starting with feature embeddings and proceeding through the PIN. Section \ref{sec:numerical} details the application of the PIN to claims frequency modeling. In Section \ref{sec:extensions}, we propose interaction importance measures for model interpretation, and we illustrate how the SHAP explanation can be computed within the PIN architecture using paired-sampling permutation SHAP. Finally, Section  \ref{sec:conclusions} concludes with a summary and future research directions.

\section{Tree-like pairwise interaction network}
\label{sec:PIN}
\subsection{Feature embedding}
We start from tabular input data $\bx=(x_1,\ldots, x_q)^\top$. In a first step, we tokenize this tabular input data to an input tensor $\bphi=\bphi(\bx)=\left[\phi_1,\ldots, \phi_q\right] \in \R^{d \times q}$; this step is done in complete analogy to Gorishniy et al.~\cite{Gorishniy}, Brauer \cite{Brauer} and
Richman et al.~\cite{RSW}. For this first step, we select a fixed embedding dimension $d \in \N$; this is a hyper-parameter selected by the modeler.

\paragraph{Categorical features.} For categorical
feature components $x_j\in{\cal X}_j=\{1,\ldots, n_j\} \subset \N$, having $n_j$ levels, consider a $d$-dimensional entity embedding $\phi_j: {\cal X}_j \to \R^d$
\begin{equation}\label{categorical entity embedding}
x_j ~\mapsto ~\phi_j(x_j) = \sum_{k=1}^{n_j} W^{(0)}_{j,k} \,\mathds{1}_{\{x_j=k\}},
\end{equation}
where $W^{(0)}_j=[W^{(0)}_{j,1}, \ldots, W^{(0)}_{j,n_j}]^\top \in \R^{n_j \times d}$ is an embedding matrix.

\paragraph{Continuous features.} For continuous feature
components $x_j \in \R$, we consider a $d$-dimensional FNN embedding $\phi_j: \R \to \R^d$
\begin{equation}\label{continuous embedding}
x_j ~\mapsto ~\phi_j(x_j) = W^{(2)}_j \,\tanh\left(W^{(1)}_j x_j + b^{(1)}_j\right) + b^{(2)}_j
\end{equation}
with biases $b^{(1)}_j \in \R^{d'}$ and $b^{(2)}_j \in \R^d$, 
weight matrices $W^{(1)}_j \in \R^{d' \times 1}$ 
and $W^{(2)}_j \in \R^{d \times d'}$, for a fixed number of 
units $d'$, and where the hyperbolic tangent function is applied element-wise.

\medskip

Collecting all embedded components $(\phi_j)_{j=1}^q=(\phi_j(x_j))_{j=1}^q$ of the input $\bx$ gives us the tensor
\begin{equation}\label{feature embedding}
\bphi=\bphi(\bx)=\left[\phi_1,\ldots, \phi_q\right]=\left[\phi_1(x_1),\ldots, \phi_q(x_q)\right] \in \R^{d \times q}.
\end{equation}
This tensor $\bphi$ involves the embedding weights $W^{(0)}_j \in \R^{n_j \times d}$ from the categorical features  (of dimension $n_jd$), and the FNN biases and weights
$(b^{(1)}_j, b^{(2)}_j , W^{(1)}_j, W^{(2)}_j)$ from the continuous feature embeddings (of dimension $2 d' + (d'+1)d$). All these biases and weights are learned from the available training data.

\subsection{Pairwise interaction layer}
The tensor $\bphi=\bphi(\bx)$, given in \eqref{feature embedding} and
containing the tokens $(\phi_j(x_j))_{j=1}^q$,
is expressing the individual feature components
of $\bx \in {\cal X}$ before they have been interacting with each other. In the next step, we are going to let these tokens interact in a pairwise manner. This can be interpreted in a similar way as the key-query interaction concept
in the attention layer of Vaswani et al.~\cite{Vaswani}.

\begin{figure}[htb!]
\begin{center}
\begin{minipage}[t]{0.45\textwidth}
\begin{center}
\includegraphics[width=\textwidth]{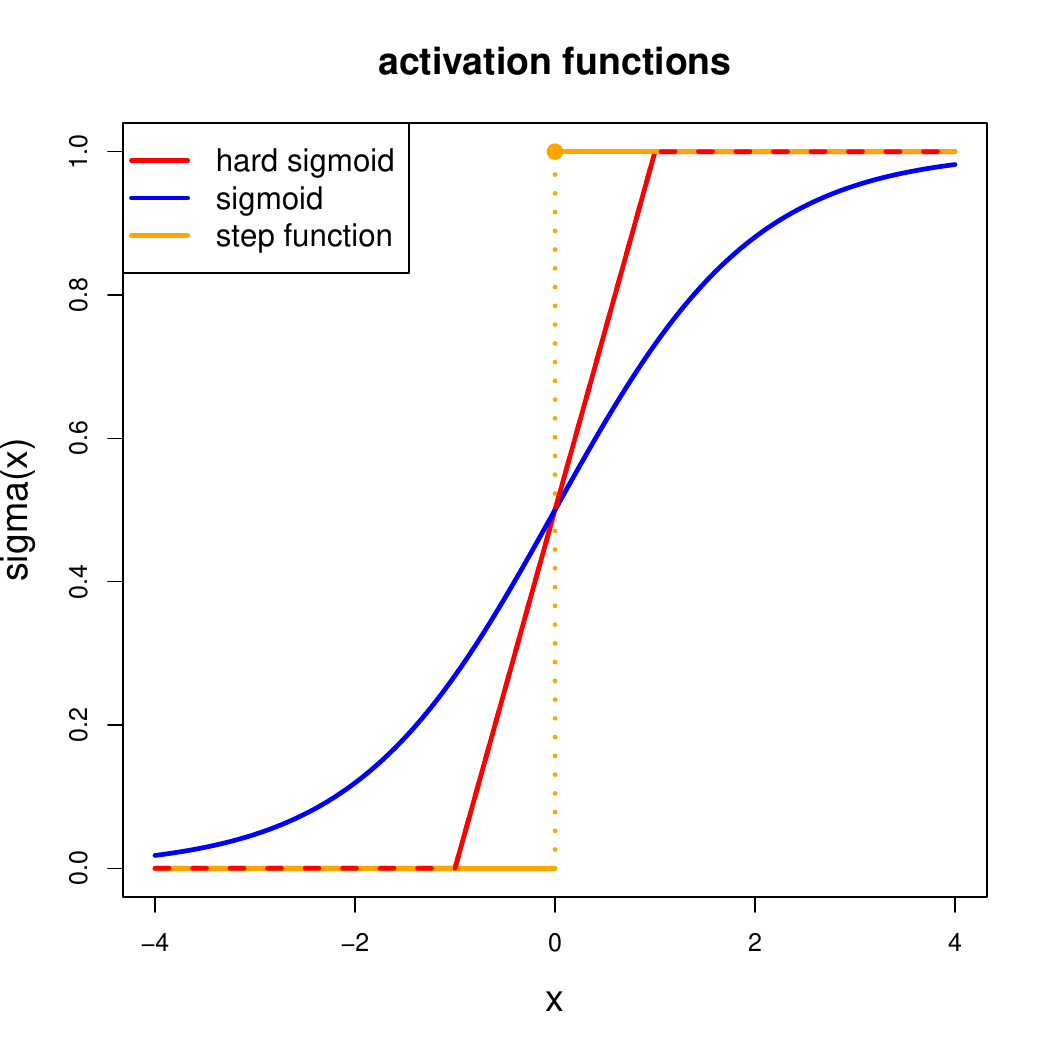}
\end{center}
\end{minipage}
\end{center}
\vspace{-.7cm}
\caption{Activation functions: (centered) hard sigmoid, sigmoid and step function.}
\label{fig: activation}
\end{figure}

We recall the (centered version of the) {\it hard sigmoid activation} 
\begin{equation}\label{hard sigmoid}
\sigma_{\rm hard}: \R \to [0,1], \qquad 
x~\mapsto~\sigma_{\rm hard}(x) 
= \max\left(0, \,\min\left(1, \, \frac{1+x}{2}\right)\right). 
\end{equation}
This is the centered version of the hard sigmoid activation;  there is also a non-centered version with $(1+x)/2$ being replaced by $x$ in \eqref{hard sigmoid}. The centered version
is directly comparable to the sigmoid activation function and the step function activation, see Figure \ref{fig: activation}.

We now let $\phi_j$ and $\phi_k$ interact, $j \le  k$.  To support these interactions we introduce {\it interaction tokens} $\be_{j,k} \in \R^{d_0}$, for a given
hyper-parameter $d_0 \in \N$. Before interacting, these interaction tokens
do not carry any information; this is similar to the CLS tokens as used
in Devlin et al.~\cite{Devlin} and Richman et al.~\cite{RSW}. We concatenate the three vectors
to a row-vector
$(\phi_j, \phi_k, \be_{j,k})^\top \in \R^{2d+d_0}$, $j \le  k$.  The goal of the
interaction tokens is to learn the interaction structure within the tensor $\bphi$ for predicting the response $Y$, in particular, 
different pairs $(\phi_j, \phi_k)$ may have a different interaction behavior which is reflected in different interaction tokens $\be_{j,k}$, $j \le  k$. Unlike the CLS tokens
of Devlin et al.~\cite{Devlin} and Richman et al.~\cite{RSW}, we do not use these interaction tokens to strip out information, but we add them to allow for the necessary modeling flexibility to be able to account for differences in the different pairs.

\medskip

Define the {\it shared interaction network}
$f_\theta: \R^{2d+d_0} \to \R$ as a composition of three FNN layers
\begin{equation}\label{shared interaction network}
f_\theta(\phi_j, \phi_k, \be_{j,k}) = W^{(3)} \,{\rm ReLU}\left(W^{(2)} \,{\rm ReLU}\left(W^{(1)} (\phi_j, \phi_k, \be_{j,k})^\top+ b^{(1)}\right) + b^{(2)}\right) + b^{(3)},
\end{equation}
with:
\begin{itemize}
\item the 1st FNN layer has $d_1$ units, rectified linear unit (ReLU) activation, 
network weights $W^{(1)}\in \R^{d_1 \times (2d+d_0)}$ and bias
 $b^{(1)}\in \R^{d_1}$;
\item the 2nd FNN layer has $d_2$ units, ReLU activation, 
network weights $W^{(2)}\in \R^{d_2 \times d_1}$ and bias
 $b^{(2)}\in \R^{d_2}$;
\item the output layer has $1$ unit, linear activation, 
network weights $W^{(3)}\in \R^{1 \times d_2}$ and bias
 $b^{(3)}\in \R$;
\end{itemize}
and $\theta=(W^{(1)},b^{(1)}, W^{(2)},b^{(2)}, W^{(3)},b^{(3)})$ collects all network parameters. This interaction network 
$f_\theta$ is going to be shared by all pairwise interactions between
the components of $\bphi$, i.e., there is one single network parameter 
$\theta$ that is shared by all ordered $1\le j \le k \le q$ pairs
$(\phi_j,\phi_k)$. 
This can also be seen as a so-called time-distributed network as being used, e.g., in recurrent neural networks. To differentiate different forms of interactions
of the different ordered pairs, $j \le k$, we let the interaction tokens $\be_{j,k}$ learn the pairwise interaction structures.

\begin{definition}[pairwise interaction layer]
For feature tokens $\phi_j(\cdot)$ and $\phi_k(\cdot)$, $1\le j \le k \le q$, and interaction tokens $\be_{j,k}$, we define the pairwise interaction units by
\begin{equation}\label{definition hjk}
h_{j,k}(\bx) 
= \sigma_{\rm hard}\Big(f_\theta\left(\phi_j(x_j), \phi_k(x_k), \be_{j,k}\right)\Big)~\in~[0,1].
\end{equation}
The pairwise interaction layer
for $\bx \in  {\cal X}$ is defined by the upper-right triangular matrix
\begin{equation}\label{pairwise interaction layer}
\left(h_{j,k}(\bx)\right)_{1\le j \le k \le q} = 
\begin{pmatrix}
h_{1,1}(\bx) & h_{1,2}(\bx) & \cdots & h_{1,q-1}(\bx)& h_{1,q}(\bx)\\
{\rm n/a} & h_{2,2}(\bx) & \cdots & h_{2,q-1}(\bx)& h_{2,q}(\bx)\\
\vdots &\vdots &\ddots &\vdots &\vdots &\\ 
{\rm n/a} & {\rm n/a} & \cdots & h_{q-1,q-1}(\bx)& h_{q-1,q}(\bx)\\
{\rm n/a} & {\rm n/a} & \cdots & {\rm n/a} & h_{q,q}(\bx)
\end{pmatrix}
,
\end{equation}
where the lower-left part is undefined indicated by {\rm n/a}.
\end{definition}

\begin{example}\normalfont
\label{example simple linear}
We give a simple example of the pairwise interaction layer.
We assume only continuous features, that is, $\bx \in \R^q$.
We select one-dimensional linear feature embeddings (tokenization) $x_j \mapsto \phi_j(x_j)=x_j$ for all $1\le j \le q$. This gives us the tensor $\bphi = \bphi(\bx)=\bx \in \R^{1 \times q}$.
For the shared interaction network $f_\theta$ we select a simple additive function and the interaction tokens are set off in this example. This gives
the simplified version
\begin{equation}\label{additive interaction example}
h^{\rm add}_{j,k}(\bx)
= 
\sigma_{\rm hard}(x_j+x_k)\qquad \text{ for $j\le k$.}
\end{equation}
This is illustrated on the left-hand side of Figure \ref{fig: interaction layer}
for $j<k$.

\begin{figure}[htb!]
\begin{center}
\begin{minipage}[t]{0.45\textwidth}
\begin{center}
\includegraphics[width=\textwidth]{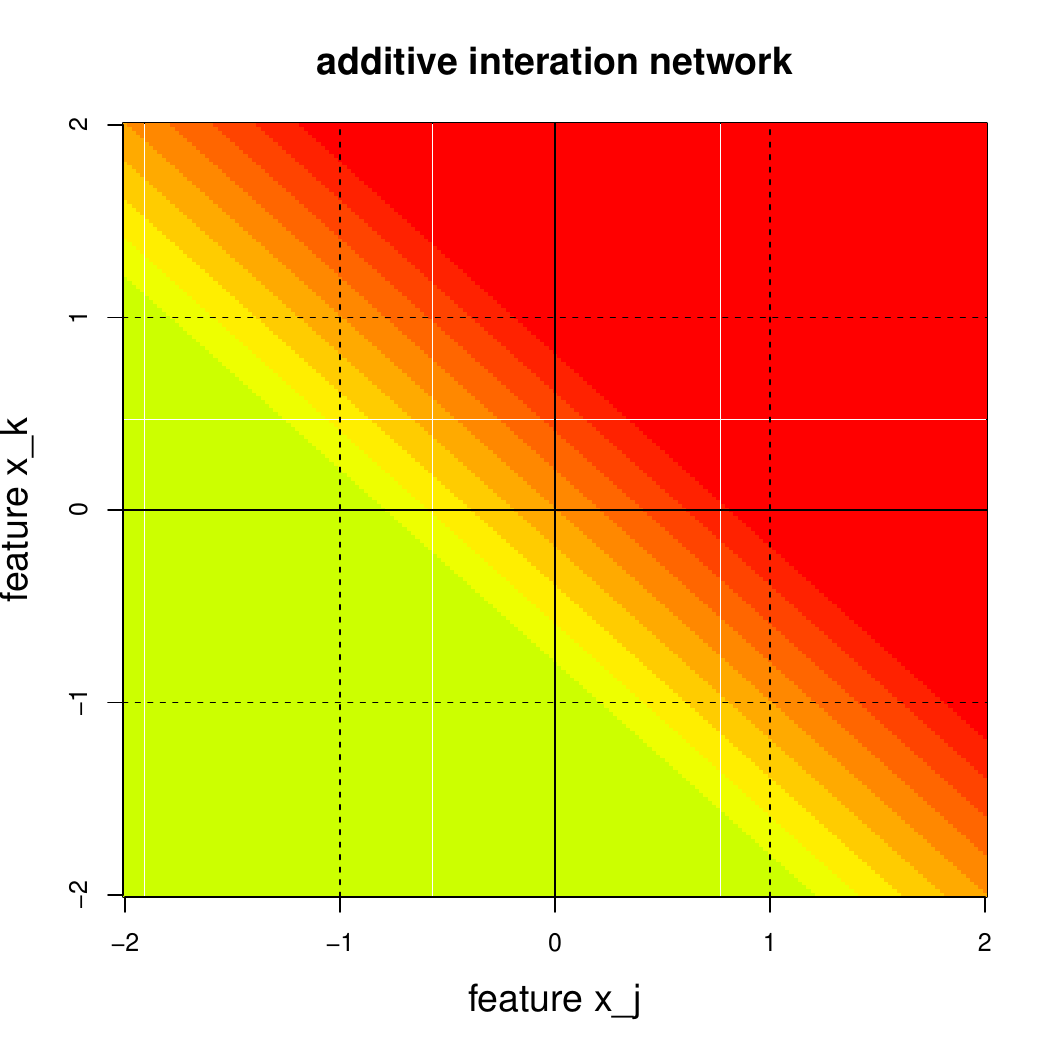}
\end{center}
\end{minipage}
\begin{minipage}[t]{0.45\textwidth}
\begin{center}
\includegraphics[width=\textwidth]{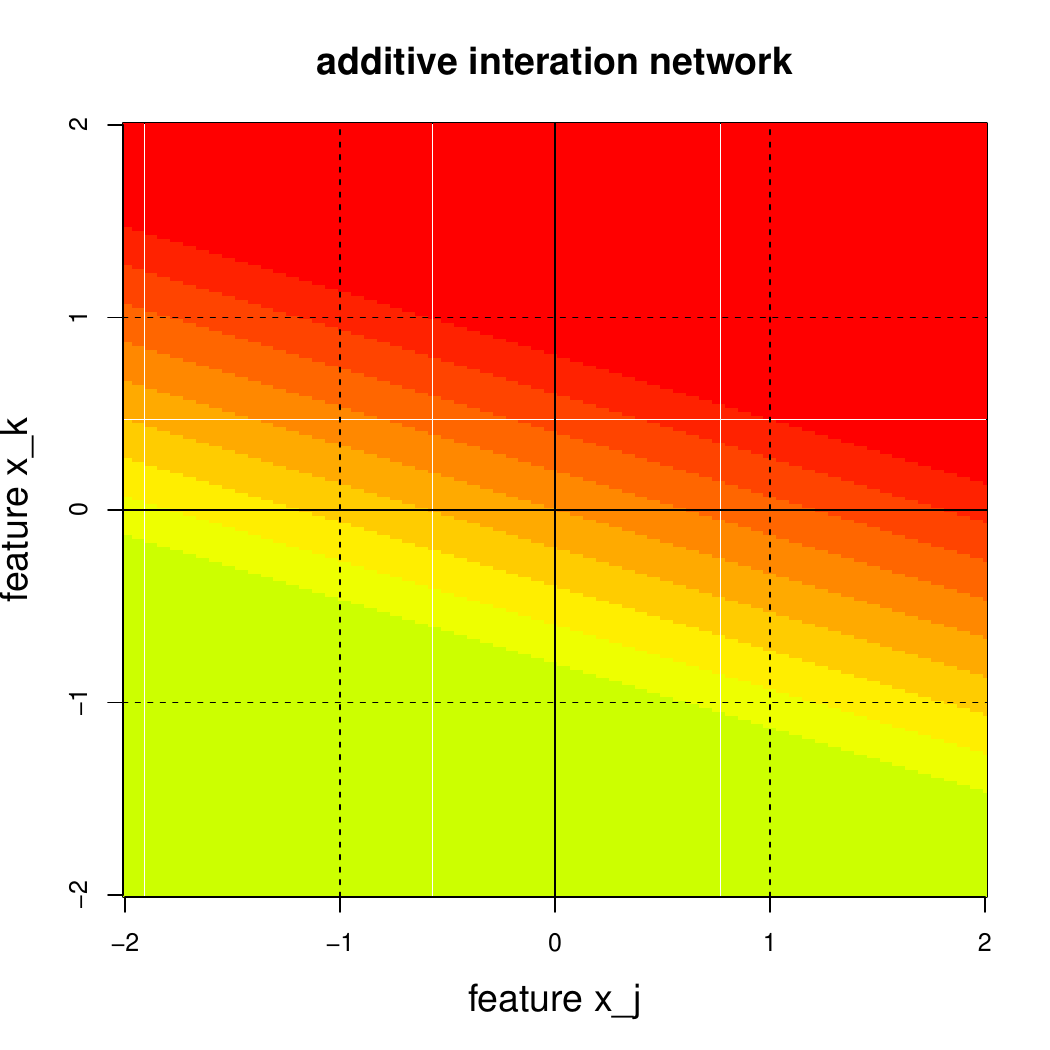}
\end{center}
\end{minipage}
\end{center}
\vspace{-.7cm}
\caption{Interaction layers: (lhs) $\sigma_{\rm hard}(x_j+x_k)$ and (rhs) $\sigma_{\rm hard}(x_j/3+x_k)$ for $j<k$.}
\label{fig: interaction layer}
\end{figure}

The right-hand side of Figure \ref{fig: interaction layer} shows a linear additive interaction network with the first component $x_j$ being scaled with $1/3$. We note that these linear additive interaction networks are rather similar to binary decision tree partitions of the feature space. We allow for additional flexibility by letting the interaction network $f_\theta$ be non-linear and by letting the interaction token $\be_{j,k}$ learn different pairwise interaction structures.
It is precisely this ability to learn to modify the inputs that gives the
pairwise interaction layers their power.
\EndExample
\end{example}

\subsection{Pairwise interaction network}
We are now ready to define the tree-like pairwise interaction network.

\begin{definition}
Choose an output activation function $g$. The pairwise interaction network (PIN)
is defined by $f_{\rm PIN}: \mathcal{X} \to \R$
\begin{equation}\label{PIN formula}
\bx ~\mapsto~
f_{\rm PIN}(\bx) = g\left(\sum_{1\le j \le k \le q} w_{j,k} \, h_{j,k}(\bx)+b\right),
\end{equation}
with learnable (real-valued) output weights $(w_{j,k})_{1\le j \le k \le q}$ and bias $b \in \R$.
\end{definition}
There is structural similarity between a PIN and a classical GBM
\begin{equation*}
f_{\rm GBM}(\bx) = \sum_{m=1}^M \sum_{l=1}^L \gamma_{m,l}\,
\mathds{1}_{R_{m,l}}(\bx), 
\end{equation*}
with $M\in \N$ GBM iterations considering binary split decision trees with $L$ leaves giving the partition $(R_{m,l})_{l=1}^L$ of the feature space ${\cal X}$ and assigning the leaf values $(\gamma_{m,l})_{l=1}^L$ to this partition. 
We start by discussing the structural correspondence
between the two regression functions.

\begin{table}[h]
\centering
\begin{tabular}{ll}
\hline
\textbf{GBM component} & \textbf{PIN analogue} \\
\hline
Indicators $\mathds{1}_{R_{m,l}}(\bx)$ & Bounded activation $\sigma_{\rm hard}(f_\theta(\cdot))$ \\
Leaf values $\gamma_{m,l}$ & Output weights $w_{j,k}$ \\
Tree structure & Network $f_\theta$ with interaction tokens $\be_{j,k}$ \\
\hline
\end{tabular}
\end{table}

We list some key differences:
\begin{enumerate}
\item \textit{Smooth activation}: The hard binary decisions 
$\mathds{1}_{R_{m,l}}(\bx)$ are replaced by continuous functions involving the hard sigmoid function.
    \item \textit{Learned representations}: There is an automatic feature transformations through embeddings resulting in the tensor $\bphi$, whereas GBMs split on the original feature values.
    \item \textit{Parameter sharing}: Common network $f_\theta$ across all interactions, and the interaction tokens learn the differences between the pairwise interactions $(\phi_j,\phi_k)$. 
    \item \textit{Joint optimization}: All components trained simultaneously, where as the GBM learns the parameters recursively in stage-wise adaptive way.
\end{enumerate}

The other close connection of the PIN is to a GA²M. A GA²M is obtained from a GAM by adding interaction terms to the regression function
\begin{equation}\label{GA2M}
g^{-1}\left(\mu(\bX)\right) = b+\sum_{j=1}^q f_j(X_j) + \sum_{1\le j<k \le q} f_{j,k}(X_j,X_k),
\end{equation}
where $f_j$ and $f_{j,k}$ are selected from a given class of splines; see Lou et al.~\cite{Lou}. A classical GAM only considers the terms $(f_j)_{j=1}^q$, and the GA²M adds the pairwise interaction terms
$(f_{j,k})_{1\le j<k \le q}$. We note that this GA²M
\eqref{GA2M} has the same structure as our PIN \eqref{PIN formula}, the difference is that we replace the splines
$(f_j)_{j=1}^q$ and $(f_{j,k})_{1\le j<k \le q}$ by pairwise interaction layers $(h_{j,k})_{1\le j\le k \le q}$ being based on FNNs.

\subsection{Connection to graph neural networks}
The pairwise interaction structure of the PIN can be naturally interpreted through the lens of graph neural networks (GNNs). We formalize this connection and demonstrate how our model can be viewed as a special case of message passing on a complete graph with learned edge features.

\begin{definition}[feature graph]
For input $\bx \in {\cal X}$, we define the corresponding feature graph $G_{\bx} = (V, E)$ as follows:
\begin{itemize}
    \item Vertices $V = \{1,\ldots, q\}$ correspond to feature components;
    \item Edges $E = \{(j,k) : 1 \leq j \le k \leq q\}$ form a complete graph;
    \item Node features $\phi_j = \phi_j(x_j) \in \R^d$ for $j \in V$;
    \item Edge features $\be_{j,k} \in \R^{d_0}$ for $(j,k) \in E$.
\end{itemize}
\end{definition}
Our PIN can be reformulated as a message passing neural network with a single update step.

\begin{definition}[message function]
The message function ${\cal M}: \R^d \times \R^d \times \R^{d_0} \to \R$ is defined as
\begin{equation*}
    {\cal M}(\phi_j, \phi_k, \be_{j,k}) = \sigma_{\rm hard}(f_\theta(\phi_j, \phi_k, \be_{j,k})),
\end{equation*}
where $f_\theta$ is the shared interaction network \eqref{shared interaction network}.
\end{definition}
It is obvious that the PIN \eqref{PIN formula} is equivalent to the GNN defined by
\begin{equation*}
    f_{\rm GNN}(\bx) = g\left(\sum_{(j,k) \in E} w_{j,k}\, {\cal M}(\phi_j, \phi_k, \be_{j,k})\right).
\end{equation*}

Our architecture differs from traditional GNNs in the following ways:
\begin{enumerate}
    \item \textit{Edge-centric}: While most GNNs focus on learning node representations, our model primarily focuses on edge interactions and the interaction token $\be_{j,k}$ to distinguish different interaction patterns.
    
    \item \textit{Learned edge features}: The interaction tokens $\be_{j,k}$ are learned parameters rather than derived features.
    
    \item \textit{Single-pass architecture}: Unlike typical GNNs that use multiple layers of message passing, our model uses a single pass with a more expressive message function.
    
    \item \textit{Bounded messages}: The hard sigmoid activation ensures bounded message values, similar to the binary nature of GBM binary decision tree splits.
\end{enumerate}
Having this connection to GNNs provides the following additional theoretical insights.

\begin{proposition}[permutation invariance]
For any permutation $\pi$ of the feature indices, the model output is invariant up to a reindexing of the output weights $w_{j,k}$.
\end{proposition}
This proposition immediately follows by analyzing permutations
of the components/vertices $V=\{1,\ldots, q\}$.

\begin{proposition}[expressiveness]
The PIN can express any message-passing GNN with
binary messages, single-layer architecture and complete graph structure.
\end{proposition}

Using GNN operations, the GNN perspective suggests several natural extensions to our base model.

\begin{definition}[multi-layer PIN]
We can define a $K$-layer model through iterative message passing, that is, 
\begin{equation}
    \phi_v^{(k+1)} = \operatorname{UPDATE}\left(\phi_v^{(k)}, \sum_{u \in {\cal N}(v)} {\cal M}^{(k)}(\phi_u^{(k)}, \phi_v^{(k)}, \be_{u,v})\right)
\end{equation}
where $\phi_v^{(0)} = \phi_v(x_v)$, $v \in V$, $k = 0,\ldots,K-1$,
and ${\cal N}(v)$ are the vertices adjacent to $v$. The $\operatorname{UPDATE}$ can be various functions such as the maximum or the mean of the two components.
\end{definition}

\begin{remark}[attention mechanism]\normalfont
The model can be enhanced with an attention mechanism by replacing the fixed output weights $w_{j,k}$ with learned functions of the node features:
\begin{equation*}
    w_{j,k} = \alpha_\vartheta(\phi_j, \phi_k),
\end{equation*}
where $\alpha_\vartheta$ is a (self-)attention layer with parameters $\vartheta$; see Vaswani et al.~\cite{Vaswani}.
If we apply this attention weights only to pairs $j<k$ 
and if we apply a thresholding, we
obtain
\begin{equation}\label{attention PIN formula}
\bx ~\mapsto~
 g\left(b+\sum_{j=1}^k w_{j} \, h_{j,j}(\bx)+\sum_{1\le j < k \le q}
{\rm ReLU}\Big(\alpha_\vartheta(\phi_j, \phi_k)-\tau\Big)\, h_{j,k}(\bx)\right),
\end{equation}
for an attention threshold $\tau \in \R$.
The thresholding in \eqref{attention PIN formula} gives a natural way of pairwise interaction selection.
\end{remark}

The connection to GNNs not only provides theoretical insights but also suggests practical extensions and implementation strategies for the PIN architecture. Moreover, for efficient implementations by existing GNN libraries are available that allow for batched processing (graph batching for parallel processing), optimized GNN kernels on GPUs, and optional sparsification of the interaction graph, similar to \eqref{attention PIN formula}.

\section{Claims frequency example}
\label{sec:numerical}
We present an example that we take in several steps to illustrate the functioning of the tree-like PIN. We consider the popular French motor third-party liability (MTPL) claims frequency dataset available from Dutang et al.~\cite{Dutang}. We apply the same data cleaning as described in W\"uthrich--Merz \cite{WM2023};\footnote{The cleaned data is available from \url{https://aitools4actuaries.com/}.} for a detailed discussion of the French MTPL dataset we refer to Appendix B of the latter reference. 

\medskip
 
Our goal is to predict the MTPL claims frequencies $Y_i=N_i/v_i$ of the insurance policyholders $i\in \{1,\ldots, n\}$, where $N_i$ is the number of claims of policyholder $i$, $v_i>0$ is the time exposure (in yearly units) of that policyholder, and $\bx_i$ describes their features. We fit different tree-like PINs to this data by using the Poisson deviance loss as the objective function; the Poisson deviance loss is the canonical strictly consistent loss function for mean estimation for claims count problems in insurance; Gneiting--Raftery \cite{GneitingRaftery}
and W\"uthrich et al.~\cite{AITools}.
For given learning data ${\cal L}=(Y_i, \bx_i, v_i)_{i=1}^n$, we aim at minimizing the following (average) Poisson deviance (PD) loss to find the optimal weights for the selected PIN 
\begin{equation}\label{Poisson deviance in-sample}
L_{\rm PD}= \frac{1}{n}
\sum_{i=1}^n 2v_i \left(f_{\rm PIN}(\bx_i)-Y_i-
Y_i \log \left(\frac{f_{\rm PIN}(\bx_i)}{Y_i}\right)\right),
\end{equation}
where $f_{\rm PIN}(\cdot)$ is the PIN given in \eqref{PIN formula}, and we select the exponential output function $g(\cdot)=\exp(\cdot)$ which is the canonical link of the Poisson model.

For model training (minimizing the Poisson deviance loss
$L_{\rm PD}$ in the network weights and biases), we use the Adam optimizer on batches of size 128, with an initial learning rate of 0.001. We then apply a learning rate reduction on plateau (factor 0.9, patience 5), and early stopping is exercised based on a 10\% validation set of the learning data ${\cal L}$; the size of the learning data is $n=610,206$. Moreover, we hold an independent test sample ${\cal T}=(Y_j, \bx_j, v_j)_{j=1}^m$ of sample size $m=67,801$, that is going to be used for out-of-sample model validation and model comparison. The learning and test samples, ${\cal L}$ and ${\cal T}$, are identical to Brauer \cite{Brauer} and Richman et al.~\cite{RSW}, which makes all results directly comparable to that reference.

\paragraph{The case of a linear PIN, with two continuous features only.}
Similar to Example \ref{example simple linear}, we start with a linear PIN, and 
we only consider the two continuous features `driver age' (DrivAge) and `log-Density'. Moreover, we only consider the interaction term $h_{1,2}(\bx)$ and we drop the diagonal terms $h_{1,1}(\bx)$ and $h_{2,2}(\bx)$ by setting $w_{1,1}=w_{2,2}=0$.
This allows us to nicely illustrate the results. For this linear off-diagonal PIN, we select the (single, $q=2$) pairwise interaction
\begin{equation*}
h^{\rm linear}_{1,2}(\bx) = \sigma_{\rm hard}\left(W^{(1)}(x_1,x_2, \be_{1,2})^\top + b^{(1)}\right).
\end{equation*}
This only considers the most inner operation of the shared
interaction network \eqref{shared interaction network}, with embedding dimension $d=1$, interaction token dimension $d_0=1$ and output dimension $d_1=1$. This gives us a weight matrix $W^{(1)}\in \R^{1 \times 3}$ and a bias $b^{(1)} \in \R$.
This is similar to the additive interaction \eqref{additive interaction example}, but additionally allowing for an affine transformation. Note that in this example of only one interaction pair, (`DrivAge', `log-Density'), the interaction token $\be_{1,2}$ would not be necessary. By inserting this linear interaction network into the PIN in \eqref{PIN formula} gives us a first illustrative example.

\begin{figure}[htb!]
\begin{center}
\begin{minipage}[t]{0.45\textwidth}
\begin{center}
\includegraphics[width=\textwidth]{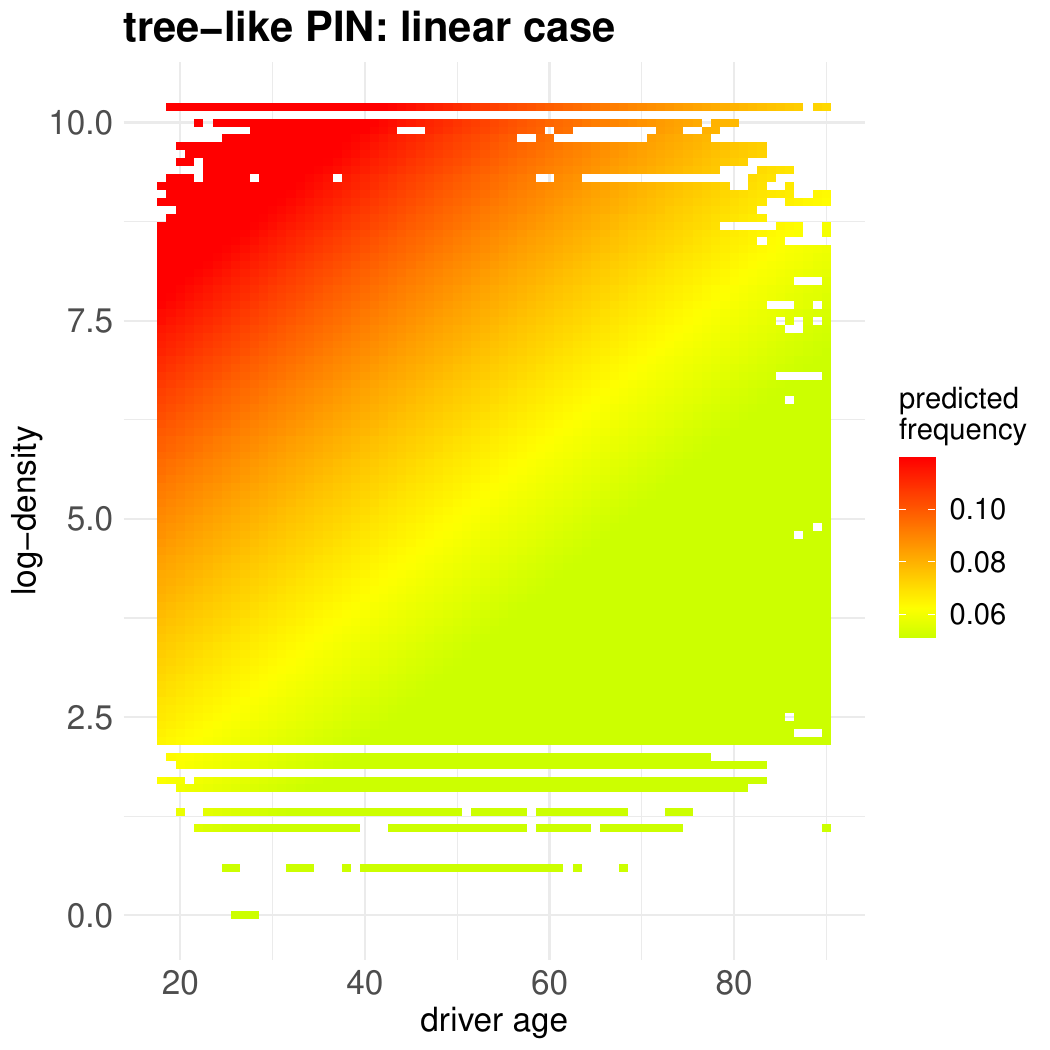}
\end{center}
\end{minipage}
\end{center}
\vspace{-.7cm}
\caption{Heatmap of predicted frequencies of the linear off-diagonal PIN only considering the features `DrivAge' and `log-Density'.}
\label{fig: PIN linear}
\end{figure}

We fit this linear off-diagonal PIN to the learning data ${\cal L}$ using the Poisson deviance loss
\eqref{Poisson deviance in-sample}, and the result is presented in Figure \ref{fig: PIN linear}. The colored dots show the estimated frequencies on the insurance portfolio (for the white parts there are no insurance policies available for these feature combinations). We observe that this 1-dimensional ($d=1$) linear off-diagonal PIN suggests a diagonal pattern. Related to decision trees, this corresponds to a feature space partition along the diagonal (for driver age and log-Density scaled as in Figure \ref{fig: PIN linear}). In the next example, we are going to refine this prediction by allowing for more general functional forms.

\paragraph{Tree-like PIN, with two continuous features only.}
In our next example, we consider the non-linear tree-like PIN as introduced in \eqref{PIN formula}. To improve the modeling step-by-step on the previous example, we again only consider the two features `DrivAge' and `log-Density' and we let them  interact non-linearly, and again we only consider the off-diagonal interaction term $h_{1,2}(\bx)$ by setting $w_{1,1}=w_{2,2}=0$. We start with the 1-dimensional embedding case $d=1$, thus, we consider 1-dimensional FNN embeddings $x_j \mapsto \phi_j(x_j) \in \R$, which we then let interact by the interaction network as defined in  \eqref{shared interaction network}; note that because we only have two features $j \in \{1, q=2\}$ in this example, this network is not shared by different interactions and the interaction token is not necessary.

\begin{figure}[htb!]
\begin{center}
\begin{minipage}[t]{0.45\textwidth}
\begin{center}
\includegraphics[width=\textwidth]{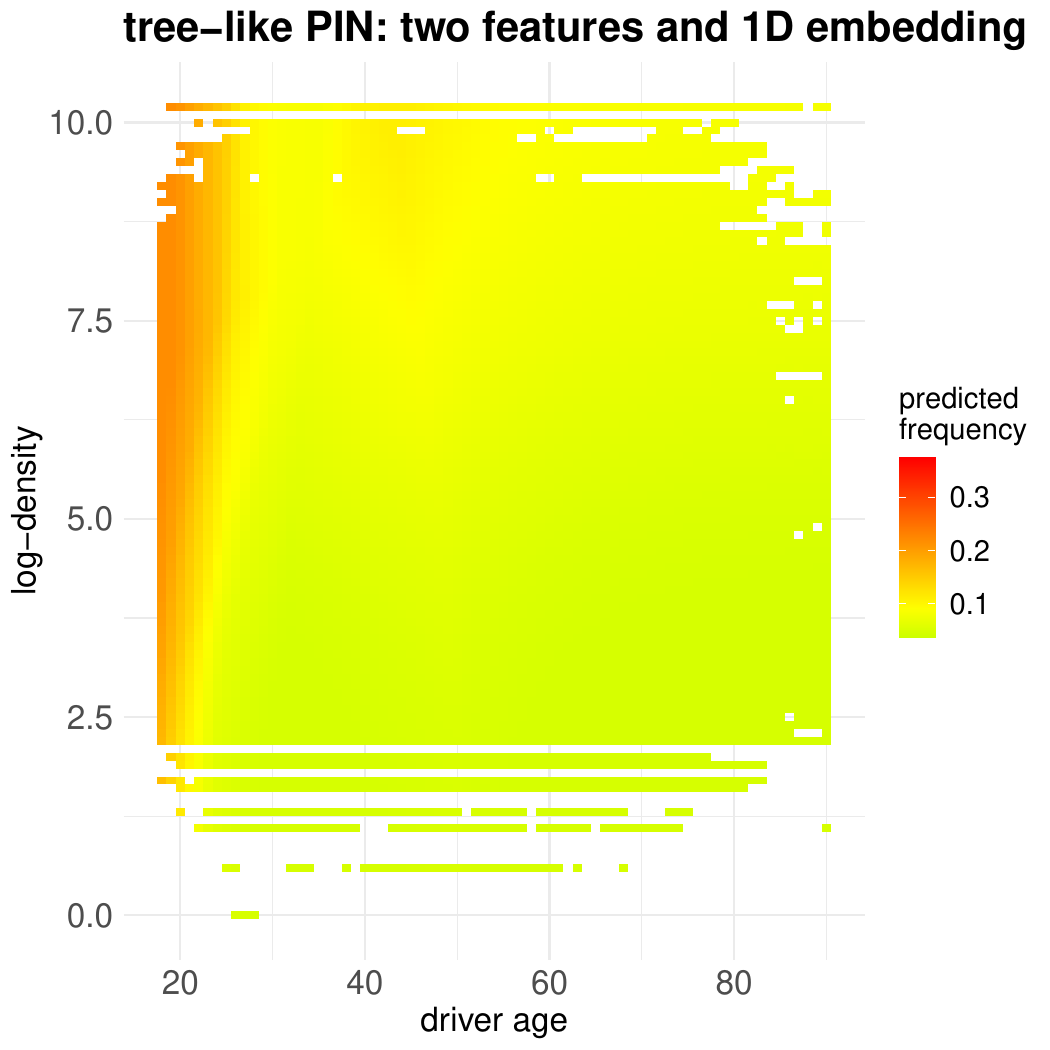}
\end{center}
\end{minipage}
\begin{minipage}[t]{0.45\textwidth}
\begin{center}
\includegraphics[width=\textwidth]{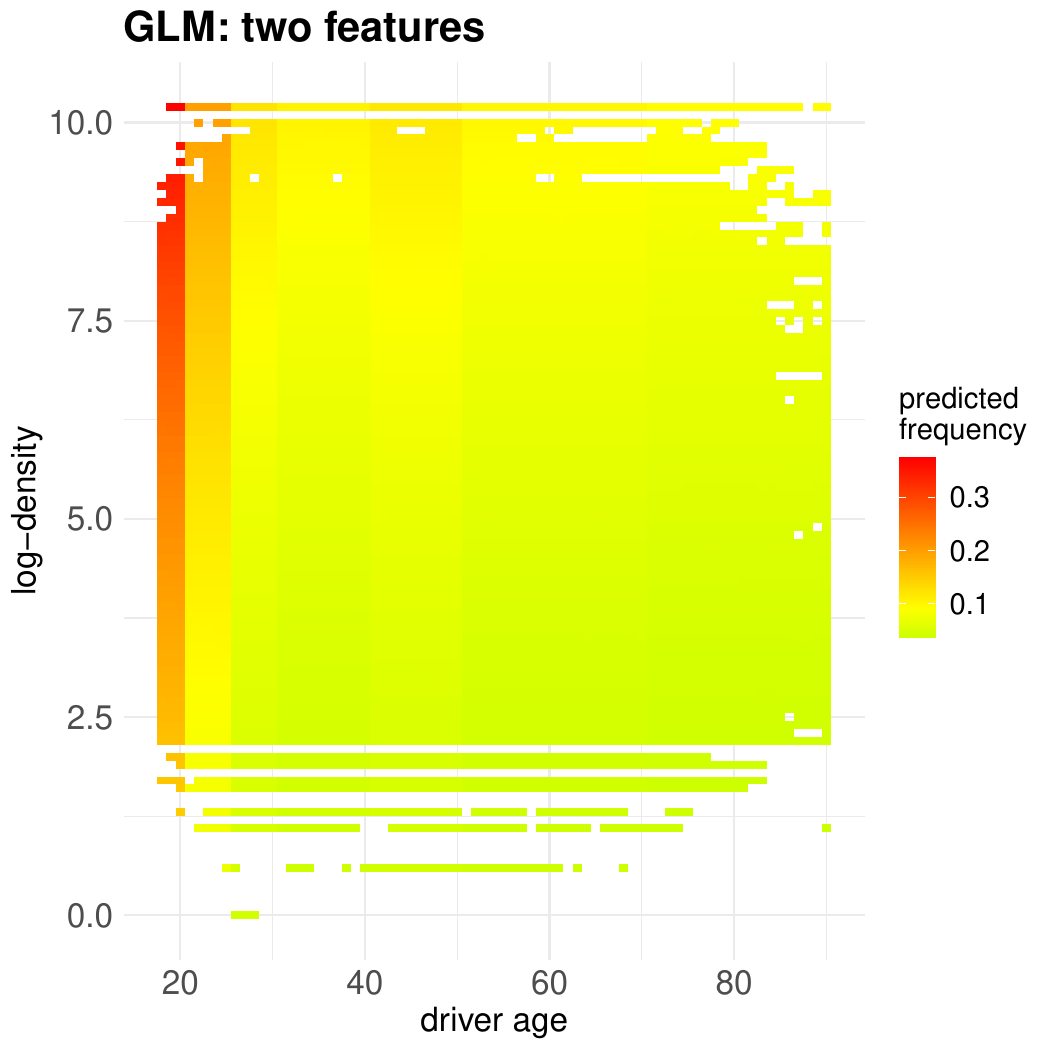}
\end{center}
\end{minipage}
\end{center}
\vspace{-.7cm}
\caption{Heatmaps of predicted frequencies of (lhs) the off-diagonal PIN with 1D embedding ($d=1$), and (rhs) GLM only considering the two features `DrivAge' and `log-Density'.}
\label{fig: PIN 2 covariates}
\end{figure}

We fit this (non-linear) off-diagonal PIN as described above by using the Adam optimizer and early stopping.
The result of this 1-dimensional case with two feature components is shown in Figure \ref{fig: PIN 2 covariates} (lhs), and it is compared to a GLM considering the same two features `DrivAge' and `log-Density', the first variable is encoded categorically and the second one continuously in this GLM, see W\"uthrich--Merz \cite[Section 5.2.4]{WM2023} for more details. The GLM in the right-hand side of 
Figure \ref{fig: PIN 2 covariates} shows a multiplicative structure in the two feature components (because of the log-link choice), and the PIN on the left-hand side gives a refinement indicating that the vertical age class coloring of the GLM is not fully appropriate. From these two plots we also observe that the expected frequency is non-monotone in the age of driver variable.

\begin{figure}[htb!]
\begin{center}
\begin{minipage}[t]{0.45\textwidth}
\begin{center}
\includegraphics[width=\textwidth]{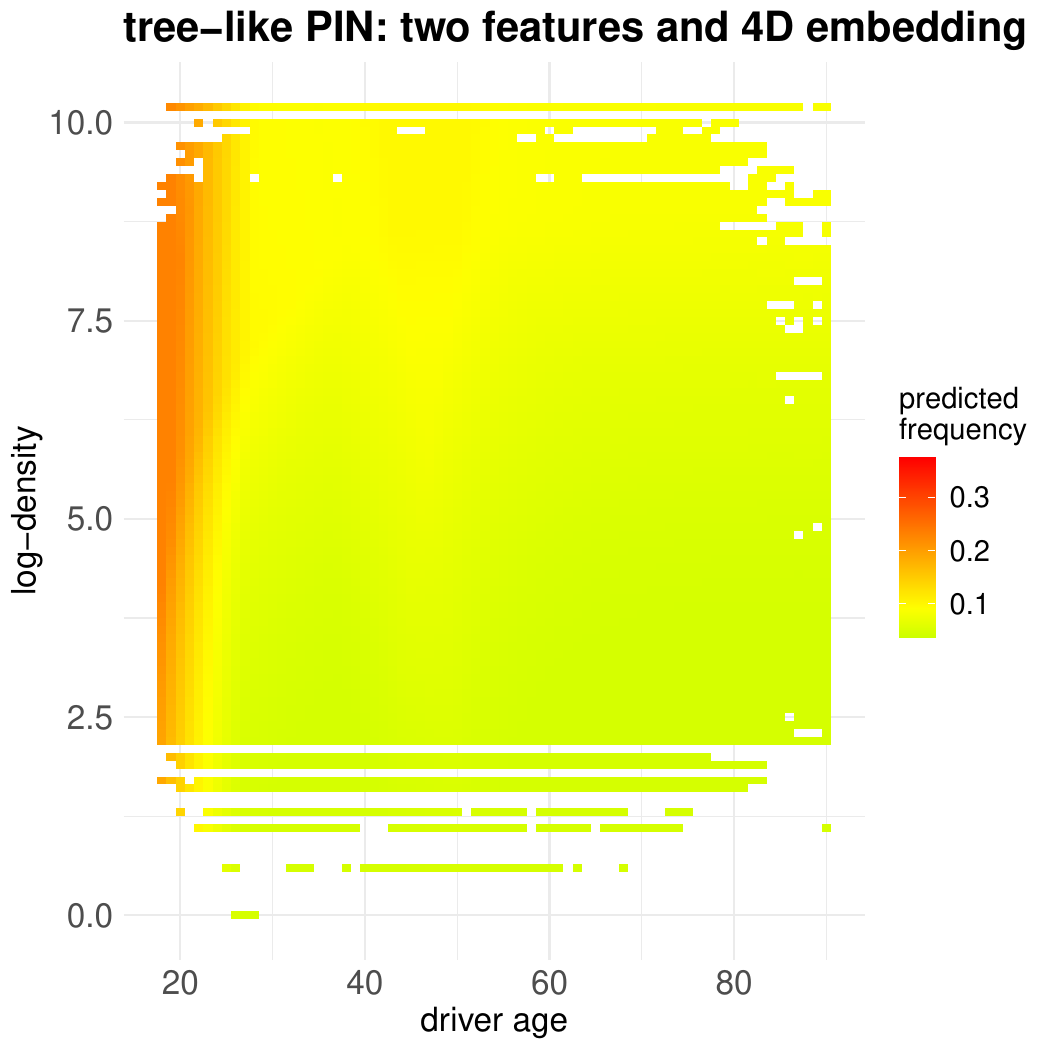}
\end{center}
\end{minipage}
\begin{minipage}[t]{0.45\textwidth}
\begin{center}
\includegraphics[width=\textwidth]{heatmap_GLM_2features.pdf}
\end{center}
\end{minipage}
\end{center}
\vspace{-.7cm}
\caption{Heatmaps of predicted frequencies of (lhs) the off-diagonal PIN with 4D embedding ($d=4$), and (rhs) GLM only considering the two features `DrivAge' and `log-Density'.}
\label{fig: PIN2 2 covariates}
\end{figure}

In a next step, a further refinement is received by increasing the dimension of the embedding to $d=4$ in $x_j \mapsto \phi_j(x_j) \in \R^d$. Similarly to GBMs, this  overlaps multiple tree-like structures. Figure \ref{fig: PIN2 2 covariates} (lhs) shows the resulting fitted off-diagonal PIN, which looks like a refinement of the GLM on the right-hand side, also more clearly reflecting the expected frequency increase in the driver age bucket between 40 and 50. Note that this age bucket has not been indicated to the PIN, but this finding of a higher frequency in this age bucket is purely data driven. Out-of-sample on the test data ${\cal T}$, the PIN has a slightly better performance than the GLM w.r.t.~the Poisson deviance loss; we do not show the figures because in this example we only showcased the functioning of the tree-like PIN on two feature components and only in the off-diagonal case ($w_{1,1}=w_{2,2}=0$); the general case is presented next.

\paragraph{Tree-like PIN on all feature components.}
We now include all available feature components and we also consider the diagonal elements of the pairwise interaction layer \eqref{pairwise interaction layer}. There are seven continuous and binary input variables to which we apply separate network embeddings \eqref{continuous embedding}, and two categorical variables to which we apply two entity embeddings \eqref{categorical entity embedding}. We select a hidden layer of dimension $d'=20$ in \eqref{continuous embedding}, and an entity embedding dimension of $d=10$. This gives us a tensor
$\bphi=\bphi(\bx) \in \R^{10 \times 9}$, and in total $10\cdot 9/2=45$ interaction tokens $\be_{j,k} \in \R^{10}$, $1\le j \le k \le 9$; we select dimension $d_0=10$ for the interaction tokens. Finally, for the shared interaction network \eqref{shared interaction network}, we select hidden layer dimensions $d_1=30$ and $d_2=20$. Table \ref{architecture table} summarizes the selected PIN architecture, indicating the number of weights/parameters to be fitted.
\begin{table}[htb!]
\centering
{\small
\begin{center}
\begin{tabular}{|l||c|r|}
\hline
Module &  $\#$ Weights
\\\hline\hline
\underline{$\bullet$ Embedding dimension $d=10$:} &\\
Continuous features (7) with $d'=20$ &  1,750 \\
Categorical features (2) with $n_j=11,22$ & 330\\
\underline{$\bullet$ Pairwise interaction layer:} &\\
Interaction tokens ($10\cdot9/2$) with $d_0=10$ &450 \\
1st FNN layer with $d_1=30$ &930 \\
2nd FNN layer with $d_2=20$ &620 \\
Output layer &21 \\
\underline{$\bullet$ Output:} &\\
Output weights $(w_{j,k})_{j\le k}$ including bias &46 \\\hline
Total &4,147 \\\hline
\end{tabular}
\end{center}}
\caption{Selected PIN architecture used on the French MTPL dataset.}
\label{architecture table}
\end{table}

This fully specifies the selected PIN architecture, which is now trained on the learning sample ${\cal L}$ using the Poisson deviance loss \eqref{Poisson deviance in-sample} as the objective function. Because early stopped stochastic gradient descent (SGD) training involves several elements of randomness, the presented results are averages over 10 SGD runs using different seeds (starting points) for the SGD algorithm, see also W\"uthrich et al.~\cite[Section 5.3.8]{AITools}. That is, the loss figures are averages over 10 different fits (and in brackets we provide the observed standard deviations across these 10 runs). Moreover, we ensemble the predictors (by averaging over the predictors). This ensembling typically improves the predictive model, which is also verified in our PIN architecture, all these results are presented in Table \ref{results PIN 1}.

\begin{table}[htb!]
\centering
{\small
\begin{center}
\begin{tabular}{|l||c|cccc|}
\hline
& \# &\multicolumn{2}{c}{In-sample\,\,}&\multicolumn{2}{c|}{Out-of-sample\,\,}\\
Model & Param.&\multicolumn{2}{c}{Poisson loss\,\,} & \multicolumn{2}{c|}{Poisson 
  loss\,\,} \\
\hline\hline
Null model (intercept-only) \cite{WM2023} &1&25.213& &25.445&\\
Poisson GLM3 \cite{WM2023}&50& 24.084 && 24.102&\\
Poisson GAM &  (66.7) &23.920&&23.956&\\\hline
Plain-vanilla FNN \cite{WM2023}&792& 23.728 & ($\pm$ 0.026)& 23.819&($\pm$ 0.017)\\
Ensemble plain-vanilla FNN \cite{WM2023}&792& 23.691 && 23.783&\\\hline
CAFFT \cite{Brauer}& 27,133& 23.715 &($\pm$ 0.047) & 23.807 &($\pm$ 0.017)\\
Ensemble CAFFT \cite{Brauer}& 27,133& 23.630 & & 23.726 &\\\hline
Credibility Transformer \cite{RSW}& 1,746& 23.641 &($\pm$ 0.053) & 23.788 &($\pm$ 0.040)\\
Ensemble Credibility Transformer \cite{RSW}& 1,746& 23.562 && 23.711 &
\\ \hline
Tree-like PIN& 4,147& 23.593 &($\pm$ 0.046) & \bl{23.740} &($\pm$ 0.025)\\
Ensemble Tree-like PIN& 4,147& 23.522 && \bl{23.667} &\\
\hline
\end{tabular}
\end{center}}
\caption{Number of parameters, in-sample and out-of-sample Poisson deviance losses (units are in $10^{-2}$); benchmark models are taken
from \cite[Table 7.9]{WM2023}, \cite[Tables 2 and 4]{Brauer} and \cite[Table 2]{RSW}.}
\label{results PIN 1}
\end{table}

Let us discuss the results of Table  \ref{results PIN 1}. The benchmark models are taken from \cite[Table 7.9]{WM2023}, \cite[Tables 2 and 4]{Brauer} and \cite[Table 2]{RSW}. The null model does not consider any features, but it is just based on the empirical mean. Poisson GLM3 is a feature-engineered GLM having 50 parameters; this model is linear on the canonical scale and does not consider any interaction terms. The linearity of the GLM is challenged by the GAM using natural cubic splines. The complexity of this GAM cannot be expressed in terms of the number of parameters involved, but the GAM routine in {\sf R} determines the effective degrees of freedom (edf) using generalized cross-validation (GCV). The resulting edf in our example are 66.7, which means that suitably chosen natural cubic splines should be able to accommodate with roughly 67 parameters. These three fitted models have been fitted by minimizing the Poisson deviance loss which is equivalent to perform MLE in the corresponding Poisson claims count model. We conclude that the GAM has the best out-of-sample performance among these three classical actuarial models, resulting in an out-of-sample loss of $23.956 \cdot 10^{-2}$. This is the best benchmark that we can get if we do not consider any interaction terms between the feature components.

If we take a deep (plain-vanilla) FNN architecture of depth 3 with $(20,15,10)$ units and with two-dimensional embedding layers for the two categorical covariates, we receive a network architecture with 792 network parameters. If we fit this architecture with an early stopped SGD with 10 initializations, we receive an average out-of-sample loss of $23.819 \cdot 10^{-2}$, and ensembling these 10 networks gives an out-of-sample performance of $23.783 \cdot 10^{-2}$, thus, interaction modeling can clearly improve the GAM.

This deep plain-vanilla FNN architecture has been improved by using different versions of the transformer architecture of Vaswani et al.~\cite{Vaswani} modified to tabular input data. The powerful adoptions of Brauer \cite[Table 4]{Brauer} and Richman et al.~\cite[Table 2]{RSW} provide out-of-sample losses of $23.726 \cdot 10^{-2}$ and $23.711 \cdot 10^{-2}$, respectively, for the ensemble versions.
From Table \ref{results PIN 1} we conclude that our tree-like PIN proposal outperforms all these other methods. Thus, this data example verifies the superior predictive performance of the tree-like PIN over some of the most recent proposals in the literature. On top, as will be illustrated in the remainder of this article, it also gives us an intrinsic explainability, in contrast to most other network architectures.

\begin{figure}[htb!]
\begin{center}
\begin{minipage}[t]{0.45\textwidth}
\begin{center}
\includegraphics[width=\textwidth]{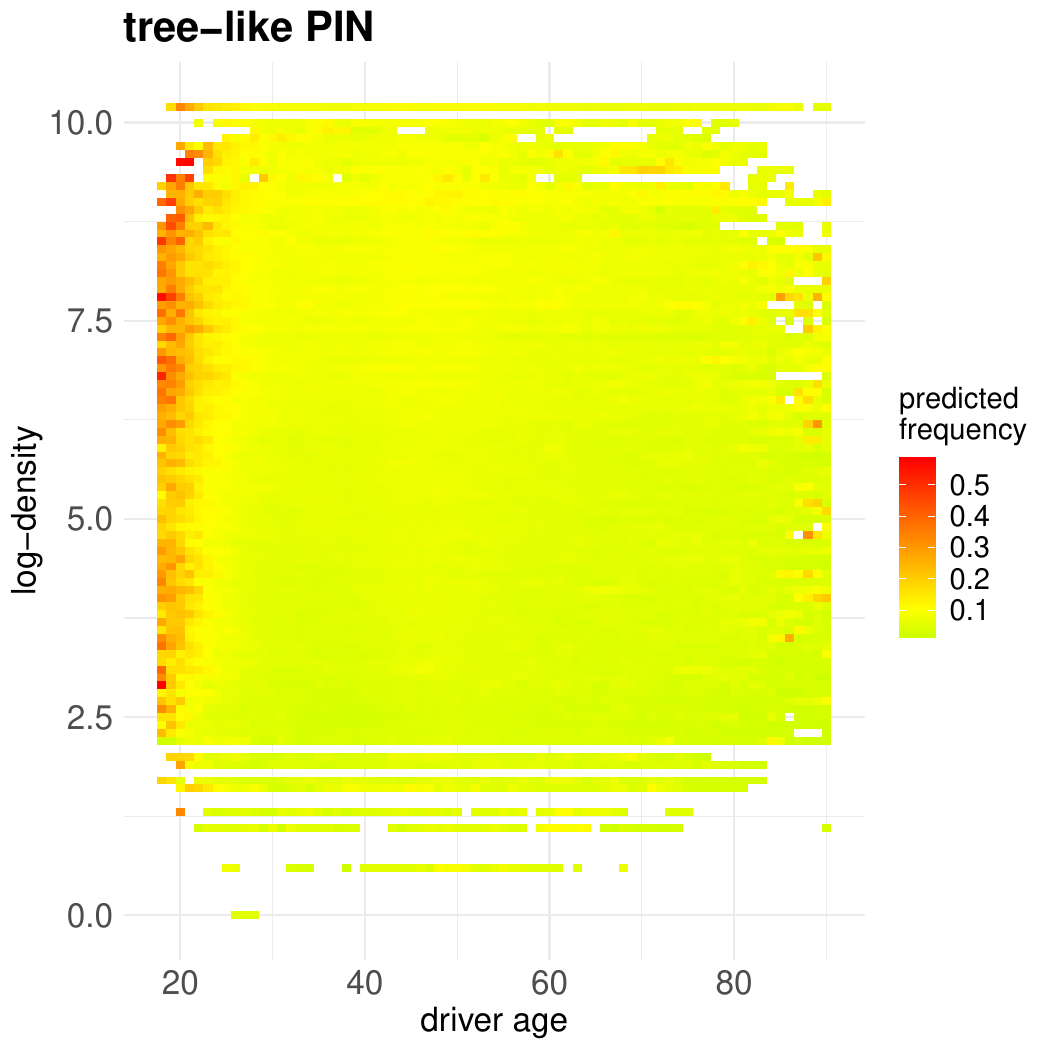}
\end{center}
\end{minipage}
\begin{minipage}[t]{0.45\textwidth}
\begin{center}
\includegraphics[width=\textwidth]{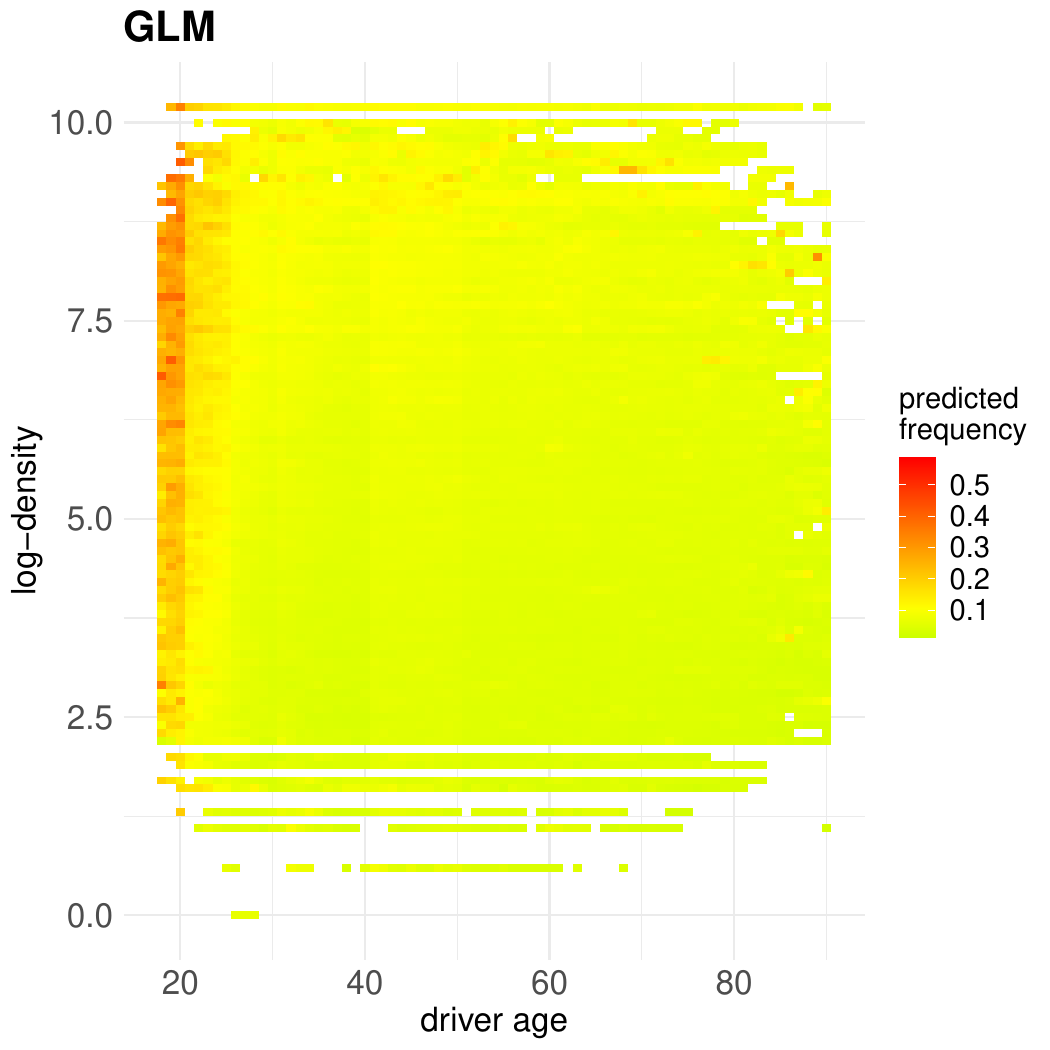}
\end{center}
\end{minipage}
\end{center}
\vspace{-.7cm}
\caption{Heatmaps of predicted frequencies of (lhs) tree-like PIN and (rhs) GLM on all available features.}
\label{fig: PIN2 all covariates}
\end{figure}

Figure \ref{fig: PIN2 all covariates} shows the resulting heatmaps of the predicted frequencies, with driver's age on the $x$-axis and log-Density on the $y$-axis; this is a two-dimensional Marginal plot (M plot); see Apley--Zhu \cite{apley2020} and W\"uthrich et al.~\cite[Section 9.3.4]{AITools}. Remark that we do not expect a smooth graph, because the predicted frequencies were computed on nine feature components, and the graphs show a projected version to two of them (M plot), i.e., these are marginalized plots. E.g., for high driver's age we see some sprinkling of red dots which can only be explained by analyzing the other features, e.g., it may be severely impacted by a high bonus-malus level; remark also that the portfolio is comparably sparse for large driver's age values, that is why extreme expected frequency values do not average out.

\section{Model explainability}
\subsection{PIN interaction importance}
\label{sec:extensions}

A notoriously difficult problem in machine learning models is explainability. Scholars have developed various post-hoc measures of variable importance, e.g., the Permutation Variable Importance (VPI) of Breiman \cite{BreimanRF} and Fisher et al.~\cite{Fisher2018}, or SHAP importance of Lundberg--Lee \cite{lundberg2017} and Mayer--Stando \cite{Michael} are two popular examples. When it comes to interactions, things become even more difficult. There is Friedman--Popescu \cite{FriedmanPopescu}'s $H^2$-statistics which is a post-hoc explainability tool that measures the impact of bivariate components on the regression function (by partial dependence functions). Apart from that there is little guidance on measuring the interaction strengths at a reasonable computational effort, i.e., testing pairwise interactions can be very costly.

Coming back to the PIN \eqref{PIN formula}, we modify this function in a way so that we do not aim at getting the maximal predictive power, but we try to evaluate which shared interaction unit $h_{j,k}(\bx)$ contributes most to a loss decrease. For this, we consider the multi-output mapping
\begin{equation}\label{multioutput PIN}
\bx ~\mapsto~
g\left(b+\sum_{l=1}^q w_{l,l} \, h_{l,l}(\bx)+
b_{j,k}+ w_{j,k} \, h_{j,k}(\bx)\right)_{1\le j < k \le q} \in~ \R^{q(q-1)/2},
\end{equation}
for biases $b_{j,k}\in \R$ and the remaining terms have the same structure as in \eqref{PIN formula}. The first part of this multi-output regression function is identical for all components $1\le j < k \le q$, that is, 
\begin{equation}\label{GAM part}
b+\sum_{l=1}^q f_l(X_l)
=b+\sum_{l=1}^q w_{l,l} \, h_{l,l}(\bx),
\end{equation}
where we make the link to the GAM \eqref{GA2M}. The remaining
terms $h_{j,k}(\bx)$ in \eqref{multioutput PIN}, $j<k$, correspond to the off-diagonal terms
of the pairwise interaction layer \eqref{pairwise interaction layer}.
Each component in \eqref{multioutput PIN}
adds another interaction term to the diagonal (non-interacting) terms \eqref{GAM part}, and we can determine the term $h_{j,k}(\bx)$, $j<k$, that leads to the biggest decrease in out-of-sample loss when added to
\eqref{GAM part}. The great thing is that we can fit all $q(q-1)/2$ interaction terms in the {\it same} gradient descent run, i.e., we apply the objective function for model fitting (simultaneously) to all components of \eqref{multioutput PIN}, and jointly fit all components by just aggregating these component-wise losses. This is possible here because we have isolated all interaction pairs in the pairwise interaction layer
\eqref{pairwise interaction layer}.

\begin{figure}[htb!]
\begin{center}
\begin{minipage}[t]{0.48\textwidth}
\begin{center}
\includegraphics[width=\textwidth]{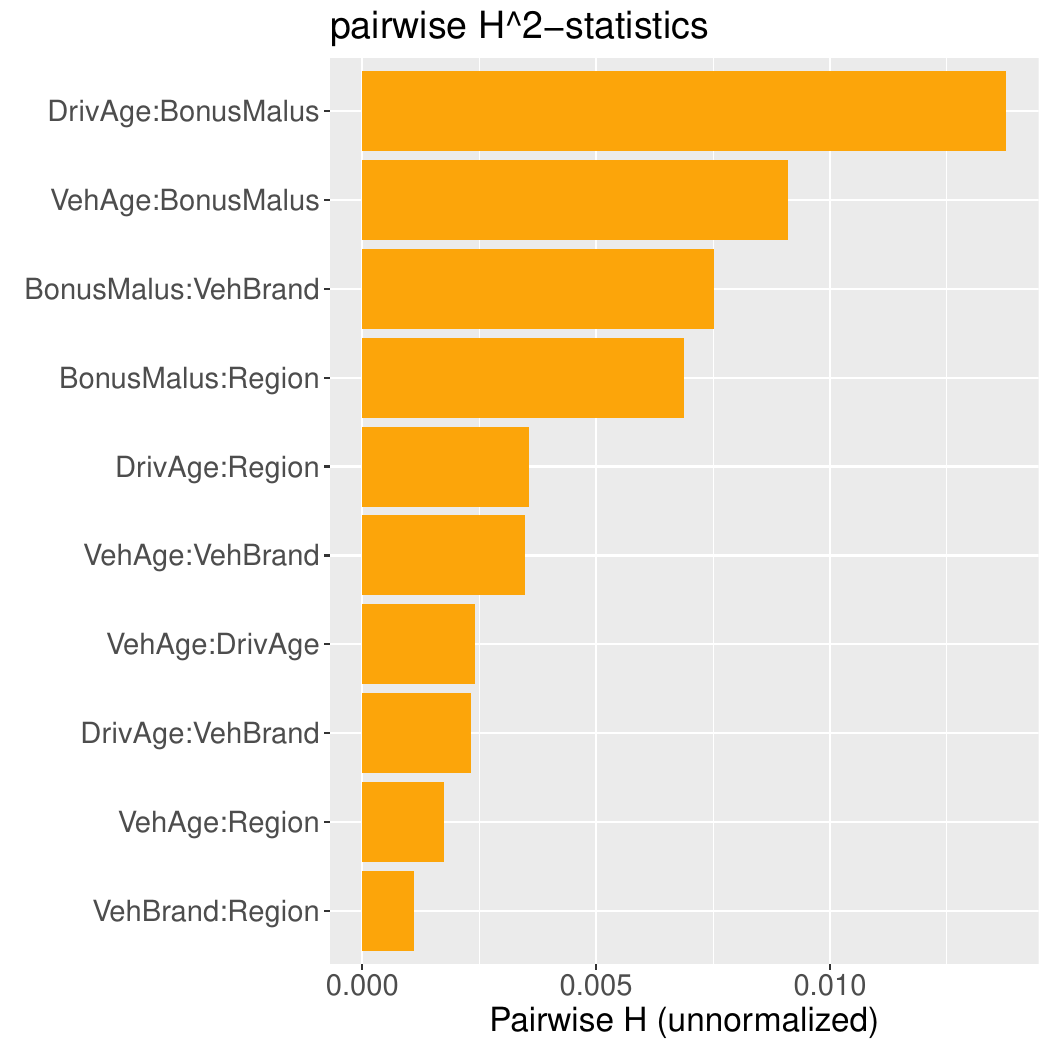}
\end{center}
\end{minipage}
\begin{minipage}[t]{0.48\textwidth}
\begin{center}
\includegraphics[width=\textwidth]{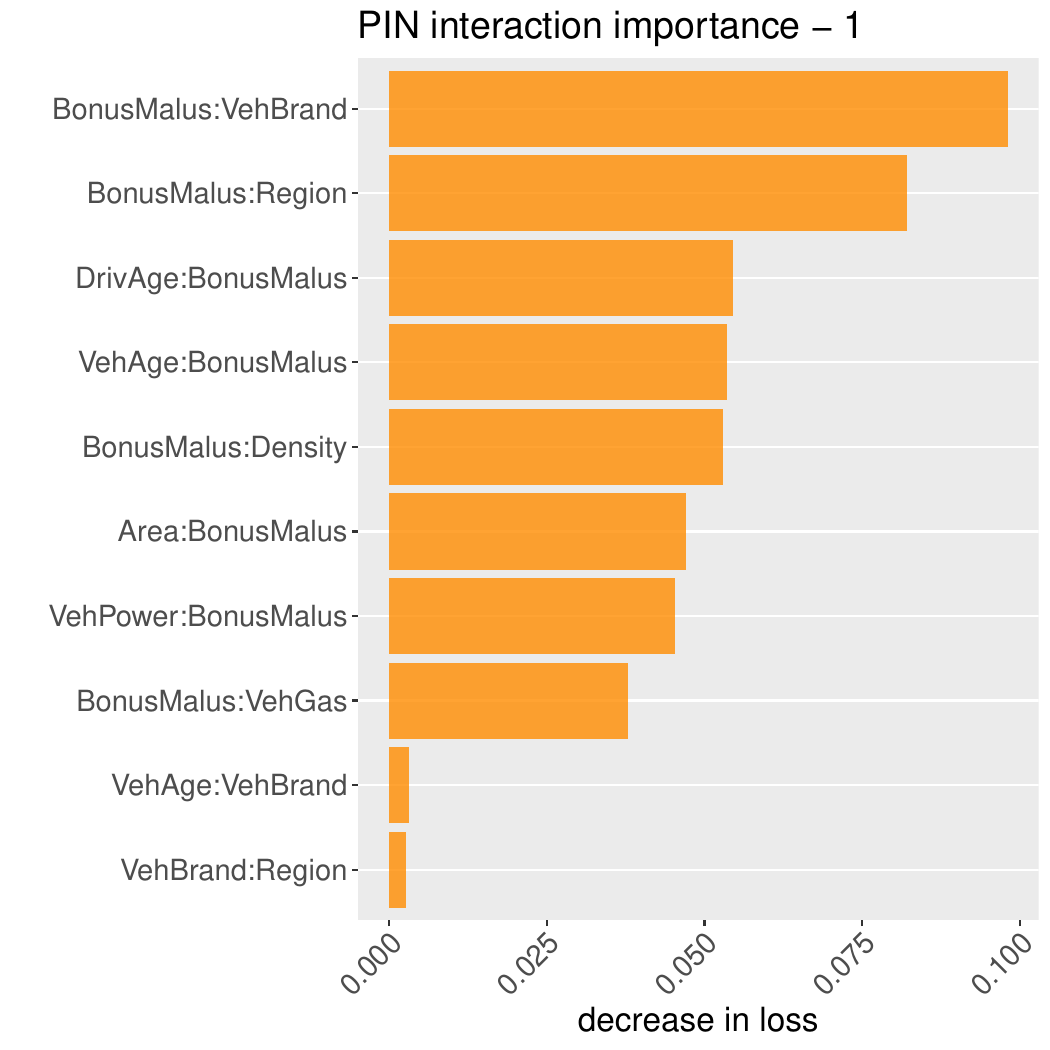}
\end{center}
\end{minipage}
\end{center}
\vspace{-.7cm}
\caption{(lhs) Friedman--Popescu \cite{FriedmanPopescu}'s $H^2$-statistics, and (rhs) PIN interaction importance.}
\label{fig: interaction statistics}
\end{figure}

In our example, we consider one further simplification, namely, we replace the diagonal (non-interacting) terms \eqref{GAM part} by the fitted GAM terms that are considered to be optimal under the exclusion of interactions, see Table \ref{results PIN 1}. This fitted GAM is kept fixed (frozen) and only the off-diagonal interaction terms in \eqref{multioutput PIN} are trained with SGD.
We then compute the decrease in out-of-sample loss compared to the GAM for each individual interaction term. Figure \ref{fig: interaction statistics} (rhs) shows the results (ranked by size). The biggest decrease in loss compared to the GAM can be achieved by adding an interaction term for (BonusMalus:VehBrand), and the second biggest decrease is obtained by adding (BonusMalus:Region).

Figure \ref{fig: interaction statistics} (lhs) illustrates
Friedman--Popescu \cite{FriedmanPopescu}'s $H^2$-statistics
of a fitted ML regression model. Not surprisingly, the same feature interactions seem to be the most impactful ones, but the order of the most important interactions is different. Friedman--Popescu's $H^2$-statistics studies the partial dependence function in a fully fitted model (which can be seen as a top-down view), whereas our PIN approach considers which interaction should be added to have the biggest impact on the prediction accuracy (which is more of a bottom-up consideration). Thus, we can interpret this as a forward selection of interaction terms, similar to the  forward selection model building in GLMs. Moreover, 
Friedman--Popescu's $H^2$-statistics is based on Partial Dependence Plots (PDP) which have the weakness that they are not reliable in cases where the feature components are highly correlated. Not surprisingly, this weakness also translates to the $H^2$-statistics. Our method does not suffer form this weakness.

\begin{figure}[htb!]
\begin{center}
\begin{minipage}[t]{0.48\textwidth}
\begin{center}
\includegraphics[width=\textwidth]{PIN_statistics.pdf}
\end{center}
\end{minipage}
\begin{minipage}[t]{0.48\textwidth}
\begin{center}
\includegraphics[width=\textwidth]{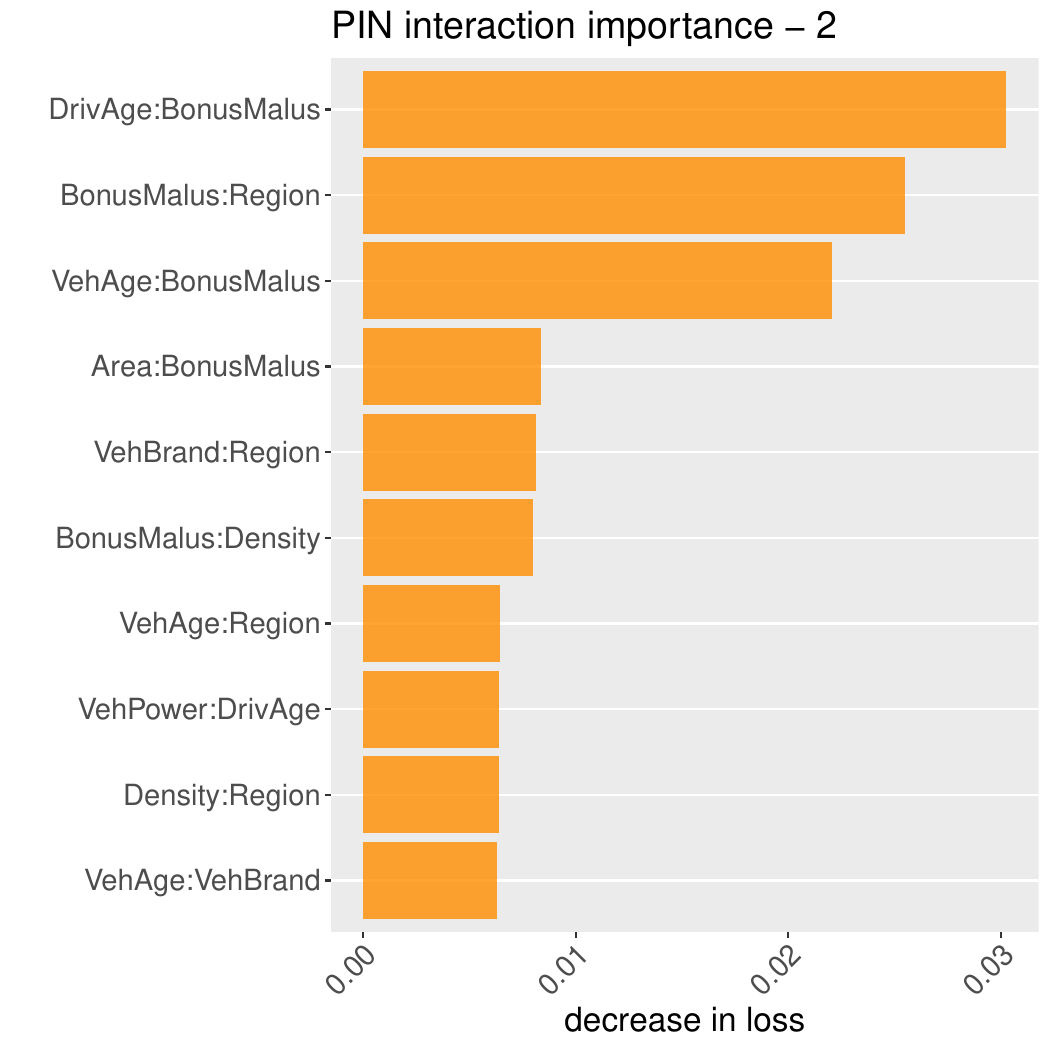}
\end{center}
\end{minipage}
\end{center}
\vspace{-.7cm}
\caption{(lhs) PIN interaction importance - 1st round (this plot is identical to Figure \ref{fig: interaction statistics}, rhs), and (rhs) PIN interaction importance - 2nd round.}
\label{fig: interaction statistics 2}
\end{figure}

Similar to the method of forward selection in GLMs, we add the most significant interaction term (BonusMalus:VehBrand) to the diagonal regression model
\eqref{GAM part}. We also freeze this interaction term, and we repeat the same fitting strategy to find the second most important interaction term, after  (BonusMalus:VehBrand) has been added to the model. The result of this second round forward selection is shown in Figure \ref{fig: interaction statistics 2} (rhs), and it is compared to the first round results on the left-hand side.
The second interaction that should be added is (DrivAge:BonusMalus).
This gives a systematic way of model building that may also be useful in improving GLMs and GAMs; this is similar to the forward selection algorithm of GLM model building. Note that this forward selected model should be refitted simultaneously on all included terms to obtain the optimal model, because in this forward inclusion process we always freeze the previously selected terms.

\subsection{SHapley's Additive exPlanation}
There is another very remarkable property of the Tree-like PIN architecture, namely, it allows for very an efficient computation of SHapley's Additive exPlanation (SHAP) as introduced by Lundberg--Lee \cite{lundberg2017}.  Shapley values originate from cooperative game theory. They are used to share a total payoff of a cooperative  game in a fair way among $q$ players; see Shapley \cite{shapley1953}. 
Denote by ${\cal C} \subseteq {\cal Q}=\{1,\ldots, q\}$ any coalition of $q$ players. Based on Shapley's fairness axioms, there is precisely one way to allocate the total payoff to the $q$ players of the cooperative game, resulting in the Shapley values $\psi_j$ assigned to players $j \in {\cal Q}$, given by
\begin{equation}\label{Shapely decomposition}
	\psi_j = \frac{1}{q} \sum_{{\cal C} \subseteq {\cal Q} \setminus\{j\}}
	{q-1 \choose |{\cal C}|}^{-1}
	\Big(\nu({\cal C} \cup \{j\}) - \nu({\cal C})\Big), \qquad \text{ for all $j \in {\cal Q}$,}
\end{equation}
for a given value function ${\cal C}\mapsto \nu({\cal C})$. In fact, starting from the given value function $\nu$, the Shapley values $(\psi_j)_{j=1}^q$ are the unique solution to a set of four axioms called Shapley's fairness axioms; see Shapley \cite{shapley1953} and Lundberg--Lee \cite{lundberg2017}.

There are two equivalent reformulations of the Shapley values \eqref{Shapely decomposition}, namely, there is the kernel SHAP formulation of 
Lundberg--Lee \cite[Theorem 2]{lundberg2017} and there is the permutation SHAP version of \v{S}trumbelj--Kononenko \cite{Strumbelj2010, Strumbelj2014}.
In this work we focus on the permutation SHAP formulation.
Denote by $\pi=(\pi_1,\ldots, \pi_q )$ a permutation of the ordered set $(1,\ldots, q)$. Let $\kappa(j) \in {\cal Q}$ be the index defined by $\pi_{\kappa(j)}=j$, and set
\begin{equation}\label{kappa function}
{\cal C}_{\pi, j} = \left\{ \pi_1,\ldots, \pi_{\kappa(j)-1}  \right\}~\subset ~{\cal Q}.
\end{equation}
These are all components of $\pi$ preceding $\pi_{\kappa(j)}=j$; for $\kappa(j)=1$, this is the empty set $\emptyset$. The permutation SHAP version of 
\v{S}trumbelj--Kononenko \cite{Strumbelj2010, Strumbelj2014} proves
that the Shapley values \eqref{Shapely decomposition} can equivalently be computed by 
\begin{equation}\label{permutation SHAP}
\psi_j = \frac{1}{q!} \sum_{\pi}\nu\left({\cal C}_{\pi, j}\cup \{j\}\right)
-\nu\left({\cal C}_{\pi, j}\right) ,
\end{equation}
where the summation runs over all permutations $\pi$. These are $q!$ permutations $\pi$, which can be an extremely large number making Shapley values computationally unattainable for large $q$. However, there is a nice result which says that if the value function $\nu$ only includes interactions of maximal order two, then an arbitrary permutation $\pi=(\pi_1,\ldots, \pi_q)$ and its reverted version
$\rho(\pi)=(\pi_q,\ldots, \pi_1)$ is sufficient to compute the Shapley values
\begin{equation}\label{permutation SHAP pair}
\psi_j = \frac{1}{2} \Big(\nu\left({\cal C}_{\pi, j}\cup \{j\}\right)
-\nu\left({\cal C}_{\pi, j}\right)
+\nu\left({\cal C}_{\rho(\pi), j}\cup \{j\}\right)
-\nu\left({\cal C}_{\rho(\pi), j}\right)\Big),
\end{equation}
if the value function $\nu$ is a bilinear form; see Mayer--W\"uthrich \cite[Proposition 3.1]{MayerW}. For example, the permutation $\pi=(1,\ldots, q)$ and its reverted version 
$\rho(\pi)=(q,\ldots, 1)$ do the job.

For explaining machine learning predictions, one typically works on the link scale, i.e., one drops the link function $g$ in \eqref{PIN formula}. For a given feature $\bx$,
this motivates the total value of the grand coalition ${\cal Q}$
\begin{equation*}
\nu({\cal Q}):=\nu_{\bx}({\cal Q}):=
g^{-1}\left(f_{\rm PIN}(\bx) \right)= \sum_{1\le j \le k \le q} w_{j,k} \, h_{j,k}(\bx)+b.
\end{equation*}
This is the prediction made by $\bx=(x_1,\ldots, x_q)^\top$, and we want to explain how the feature components $x_j$ contribute to this prediction. This requires the choice of the value function $\nu({\cal C})=\nu_{\bx}({\cal C})$ for any coalition ${\cal C}\subset {\cal Q}$ if the elements in ${\cal Q}\setminus {\cal C}$ are masked (not available for prediction). The most popular choice for describing the masking is the so-called interventional SHAP version, in our case given by
\begin{equation}\label{interventional SHAP}
\nu({\cal C}):=\nu_{\bx}({\cal C}) := \E\left[\sum_{1\le j \le k \le q} w_{j,k} \, h_{j,k}\left(\bx_{\cal C}, \bX_{{\cal Q}\setminus{\cal C}}\right)+b\right],
\end{equation}
where the components $\bx_{\cal C}=(x_j)_{j \in {\cal C}}$ are kept fixed and the components
$\bX_{{\cal Q}\setminus{\cal C}}=(X_j)_{j \in {\cal Q}\setminus{\cal C}}$ are resampled from the population distribution
$\bX \sim \p$. An important point in the subsequent computation of the Shapley values is that each term $h_{j,k}$ in \eqref{interventional SHAP} only depends on the components $(j,k)$, i.e., we only have pairwise interactions, see
\eqref{definition hjk}. That is, by a slight abuse of notation, 
\begin{equation*}
h_{j,k}\left(\bx_{\cal C}, \bX_{{\cal Q}\setminus{\cal C}}\right) = \left\{
    \begin{array}{ll}
      h_{j,k}(x_j, x_k) & \text{ if $j,k \in {\cal C}$,}\\
      h_{j,k}(X_j, x_k) & \text{ if $j \in {\cal Q} \setminus {\cal C}$ and $k \in {\cal C}$,}\\
      h_{j,k}(x_j, X_k) & \text{ if $j \in {\cal C}$ and $k\in {\cal Q} \setminus {\cal C}$,}\\
      h_{j,k}(X_j, X_k) & \text{ if $j,k \in {\cal Q} \setminus {\cal C}$.}\\
    \end{array}
  \right.
  \end{equation*}
This interventional SHAP formula \eqref{interventional SHAP} is usually estimated empirically. For this we select a background set ${\cal B}$ from the learning sample ${\cal L}$. This allows one to determine the value function empirically by
\begin{equation}\label{interventional SHAP empirical}
\widehat{\nu}({\cal C})=\widehat{\nu}_{\bx}({\cal C}) = 
\frac{1}{|{\cal B}|}\sum_{i \in {\cal B}}
\left[\sum_{1\le j \le k \le q} w_{j,k} \, h_{j,k}\left(\bx_{\cal C}, \bX^{(i)}_{{\cal Q}\setminus{\cal C}}\right)+b\right],
\end{equation}
where $\bX^{(i)}$ is the $i$-th instance of the background dataset.

The crucial point now is that the value function ${\cal C} \mapsto \widehat{\nu}({\cal C})=\widehat{\nu}_{\bx}({\cal C})$ only involves interactions of maximal order two. Therefore, the
paired permutation SHAP formula \eqref{permutation SHAP pair} applies (and is exact for the selected value function $\widehat{\nu}=\widehat{\nu}_{\bx}$). This requires $2(q+1)$ evaluations of the value function $\widehat{\nu}$, and each value function evaluation requires $|{\cal B}|$ computations of  PIN predictions, see \eqref{interventional SHAP empirical}. Thus, altogether the SHAP decomposition of the prediction $g^{-1}(f_{\rm PIN}(\bx) )$ requires
$2(q+1)|{\cal B}|$ PIN network computations for computing the {\it exact} Shapley values $(\psi_j)_{j=1}^q$ of the value function $\widehat{\nu}$. Any other network architecture that involves higher order interactions would need $q!\, |{\cal B}|$  network computations for receiving the exact Shapley values of the corresponding estimated interventional SHAP value function, thus, the order of computations grows as $q!$ instead of $2(q+1)$ as in our Tree-like PIN case.

\medskip

We revisit the PIN architecture depicted in Table \ref{architecture table}
having the best predictive performance of all models presented in Table \ref{results PIN 1}. We emphasize that this network architecture only considers interactions of order two on the link scale, thus, the paired-sampling permutation SHAP formula
\eqref{permutation SHAP pair} applies, and it gives exact results, for instance, we can use the permutation pair $\pi=(1,\ldots, q)$ and $\rho(\pi)=(q,\ldots, 1)$. As background dataset we select at random $|{\cal B}|=2000$ instances from the learning set. This allows us to decompose the predictions $g^{-1}(f_{\rm PIN}(\bx) )$ of selected
instances $\bx$. Decomposing 100 predictions takes 23 seconds on an ordinary laptop.\footnote{Intel Core i7-1355U Processor (12M Cache, up to 5.00 GHz)}

\begin{figure}[htb!]
\begin{center}
\begin{minipage}[t]{0.48\textwidth}
\begin{center}
\includegraphics[width=\textwidth]{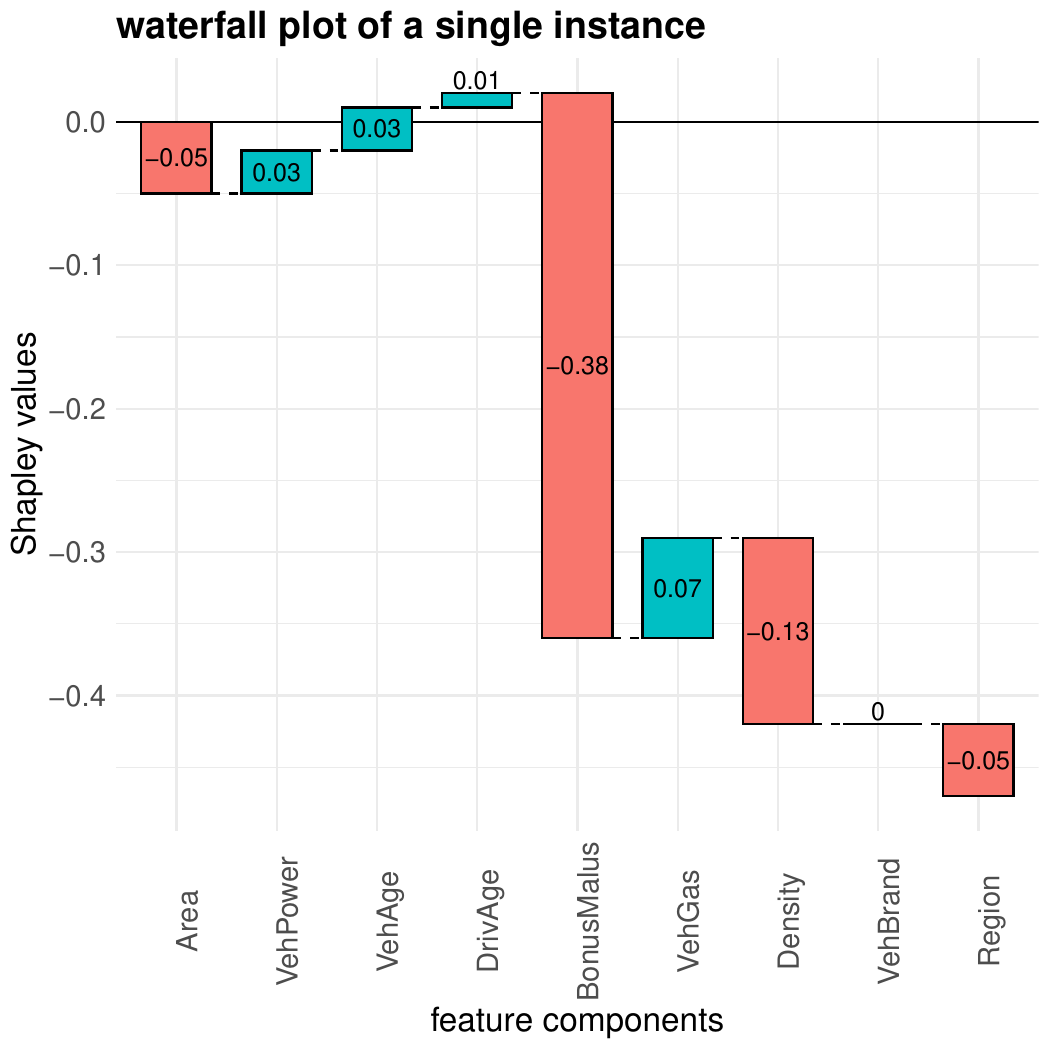}
\end{center}
\end{minipage}
\begin{minipage}[t]{0.48\textwidth}
\begin{center}
\includegraphics[width=\textwidth]{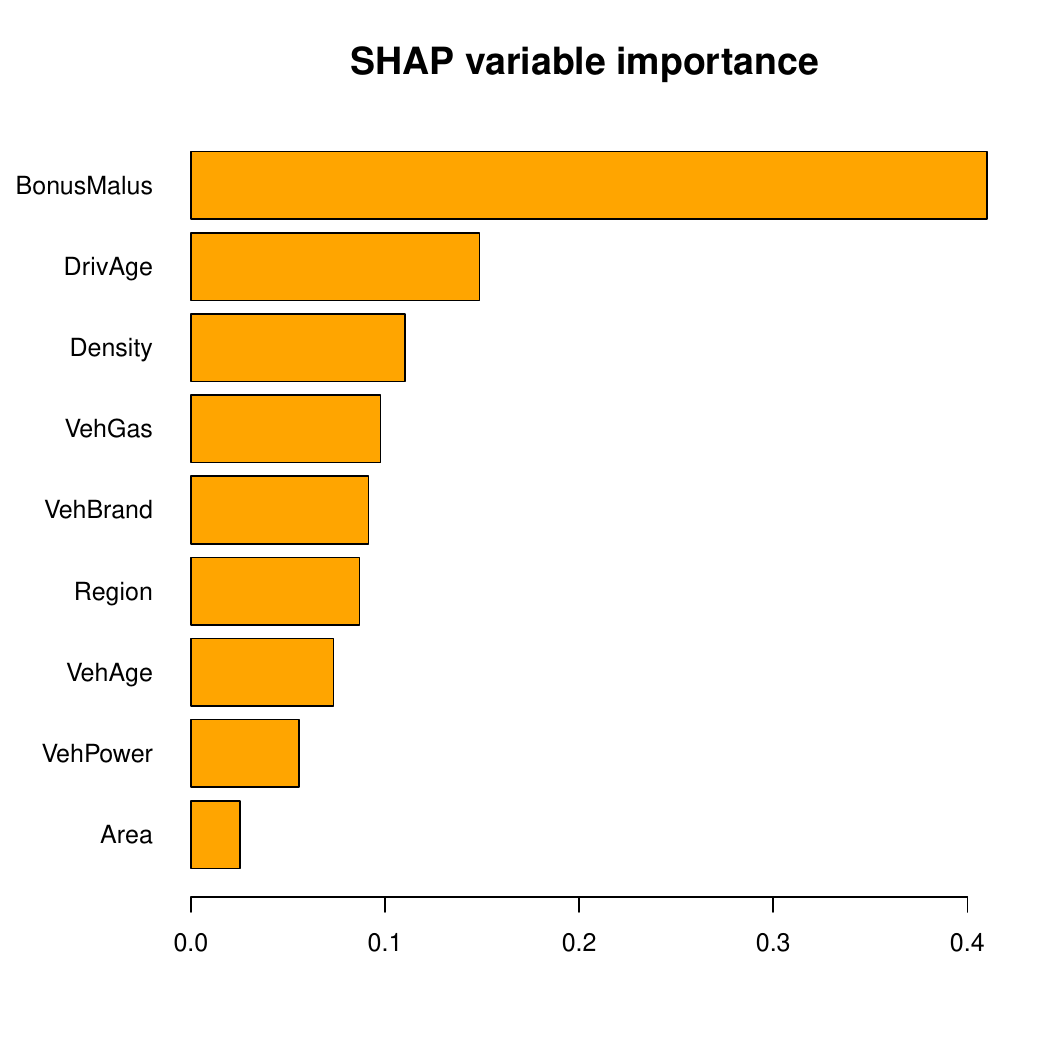}
\end{center}
\end{minipage}
\end{center}
\vspace{-.7cm}
\caption{(lhs) SHAP waterfall graph $(\psi_j)_{j=1}^q$ of a single instance $\bx$, and (rhs) SHAP variable importance.}
\label{fig: SHAP 1}
\end{figure}

The left-hand side of Figure \ref{fig: SHAP 1} shows the Shapley values $(\psi_j)_{j=1}^q$ of a single instance $\bx$. We have global mean of the background dataset of $\widehat{\nu}_{\bx}(\emptyset)=-2,776$, and the prediction of feature $\bx$ is given by $\widehat{\nu}_{\bx}({\cal Q})=g^{-1}(f_{\rm PIN}(\bx) )=-3.239$. Thus, feature $\bx$ has a prediction being $-0.463$ below average. The Shapley values $(\psi_j)_{j=1}^q$ illustrated in Figure \ref{fig: SHAP 1} (lhs) show how this lower prediction can be explained. The biggest decrease is explained by the bonus-malus level and the density features, whereas the vehicle gas feature leads to the biggest increase in the explanation of this difference from $\widehat{\nu}_{\bx}(\emptyset)$ to $\widehat{\nu}_{\bx}({\cal Q})$.

Going from local to global SHAP explanations, we can decompose the predictions $g^{-1}(f_{\rm PIN}(\bx_i))$ of many instances $(\bx_i)_{i \in {\cal I}}$, for a randomly selected subset ${\cal I}$ of the entire data. This provides us with SHAP decompositions $(\psi^{(i)}_j)_{j=1}^q$ which allow us to consider different statistics. The following analysis is based on $|{\cal I}|=1000$ instances $\bx_i$ that have been decomposed by the paired-sampling permutation SHAP method. SHAP variable importance is then obtained by considering
\begin{equation*}
\overline{\psi}_j = \frac{1}{|{\cal I}|} \sum_{i \in {\cal I}} |\psi^{(i)}_j|.
\end{equation*}
Figure \ref{fig: SHAP 1} (rhs) shows these importance measures $\overline{\psi}_j$ for all feature components $j \in {\cal Q}$. The most significant role is played by the bonus-malus level (past claims history) followed by the driver's age variable, at the other end is the area code which seems to be the least important variable.

\begin{figure}[htb!]
\begin{center}
\begin{minipage}[t]{0.48\textwidth}
\begin{center}
\includegraphics[width=\textwidth]{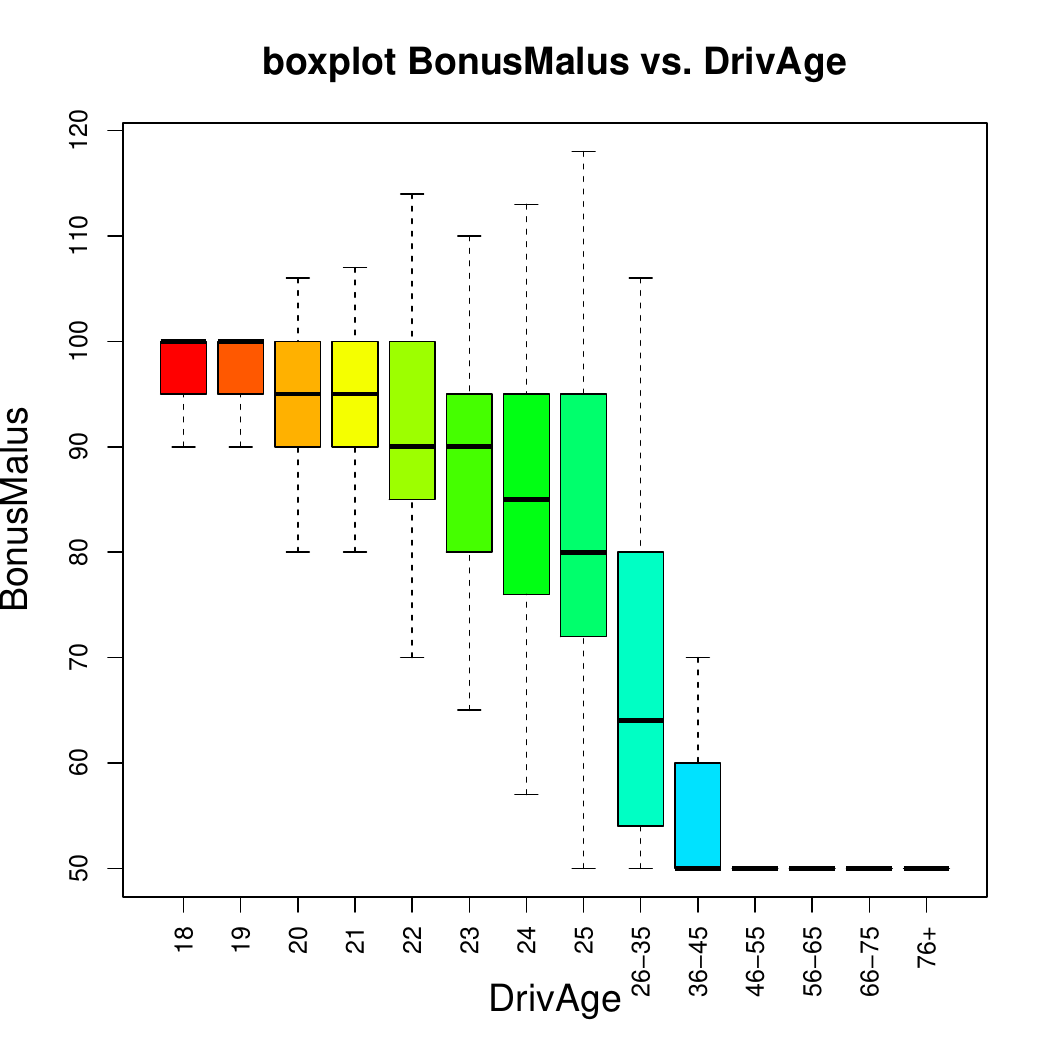}
\end{center}
\end{minipage}
\begin{minipage}[t]{0.48\textwidth}
\begin{center}
\includegraphics[width=\textwidth]{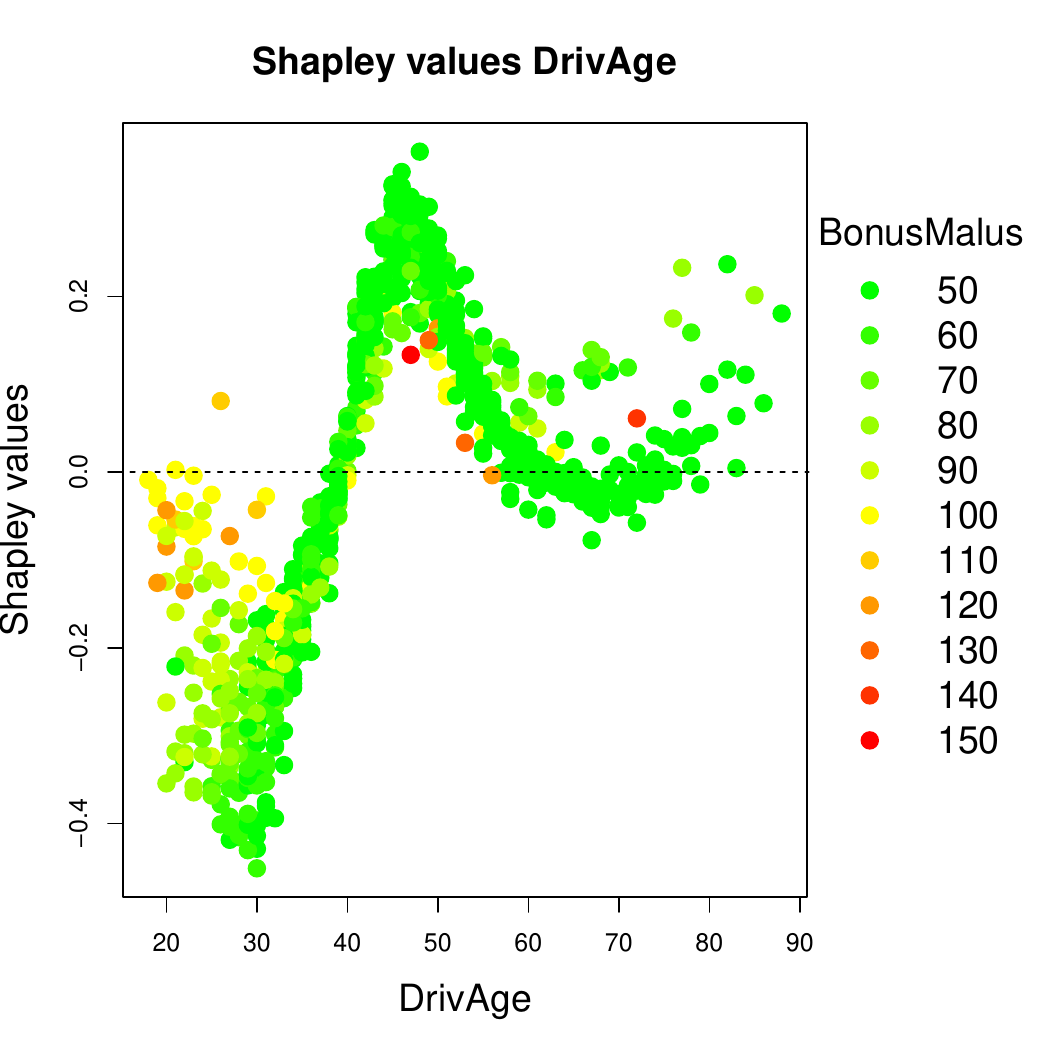}
\end{center}
\end{minipage}
\end{center}
\vspace{-.7cm}
\caption{(lhs)  Dependence between driver's age and the bonus-malus variable, this graph is taken from \cite[Figure 13.12]{WM2023}, and (rhs) Shapley values $\psi_j^{(i)}$ of the selected 100 instances $i\in {\cal I}$ for the covariate $j={\tt DrivAge}$; the coloring is the bonus-malus level of these instances.}
\label{fig: SHAP 2}
\end{figure}

In Figure \ref{fig: SHAP 2} (rhs), we analyze the 1000 decomposed instances $\bx_i$, $i\in {\cal I}$. The plot shows the Shapley values $\psi^{(i)}_j$ of these 1000 instances for the variable $j ={\tt DrivAge}$. This driver age variable has its strongest interaction with the bonus-malus level, see Figure \ref{fig: interaction statistics 2}, and the coloring of the Shapley values in Figure \ref{fig: SHAP 2} (rhs) is selected according to these bonus-malus levels. We observe that the driver's age variable has a decreasing effect on the prediction for drivers' ages below 40 (negative Shapley values) and an increasing effect for ages between 40 and 60 (positive Shapley values). To fully understand this picture, one needs to understand the interaction with the bonus-malus level, and we observe for drivers' ages below 40, that this positive age effect especially holds for low bonus-malus levels. In fact, there is a strong dependence between the driver's age variable and the bonus-malus level for drivers' ages below 40, see boxplot of Figure \ref{fig: SHAP 2} (lhs), because new drivers start at a bonus-malus level of 100 (yellow color in Figure \ref{fig: SHAP 2}, rhs). This level decreases in every year that the driver is free of accidents, and otherwise it increases. Therefore, young drivers typically have a higher bonus-malus level than mature drivers (in a low frequency environment), and the Shapley values in Figure \ref{fig: SHAP 2} (rhs) tell us that young drivers with a lower bonus-malus level (green colors) get a decrease compared to their older peers in the same bonus-malus class. Of course, this interaction makes perfect sense. This completes our example.

\section{Conclusion}
\label{sec:conclusions}

This paper introduces the Tree-like Pairwise Interaction Network (PIN), a novel neural  network architecture designed to explicitly capture pairwise covariate interactions in tabular data. Modeling such interactions is essential in predictive tasks, as they reveal how covariates jointly affect the response beyond their individual contributions. The Tree-like PIN embeds each input covariate into a latent space, and it models pairwise interactions through a shared feed-forward neural network equipped with a centered hard sigmoid activation, which replicates the partitioning behavior of decision trees. The use of a shared network ensures parameter efficiency, while the inclusion of interaction-specific learnable parameters preserves the flexibility required to capture heterogeneous dependencies across covariate pairs.

Numerical experiments show that the Tree-like PIN achieves a strong predictive performance and it consistently outperforms both traditional benchmarks and recently introduced neural network architectures, including the Credibility Transformer of Richman et al.~\cite{RSW}. In addition, PIN enhances model interpretability by enabling the analysis of how pairs of covariates jointly contribute to the response. Since this architecture only involves pairwise interactions, it allows for a very efficient way of computing SHapley's Additive exPlanations (SHAP) by the paired-sampling permutation SHAP method.
The paper also highlights connections between the functioning of the PIN architecture and existing methods. Despite its novel architecture, PIN shares similarities with GA²Ms, gradient boosting, and graph neural networks.

Possible future research directions include the incorporation of regularization techniques to promote sparsity, thereby enabling an automated selection of the most relevant interaction pairs and improving model generalization. Such regularization strategies could also be employed to penalize terms involving protected attributes, supporting the development of fairer predictive models. Another promising avenue is the extension of the PIN architecture to time series data, enabling the joint modeling of temporal dependencies and feature interactions.

\bigskip

{\bf Conflict of interest.} The authors are not aware of any conflict of interest.

\medskip

{\bf Code and data availability.} Code and data is available from the GitHub repository:\\
  https://github.com/wueth/Tree-Like-PIN

\bibliographystyle{plain}

\end{document}